\documentclass{article}
\makeatletter
\newcommand{\dontusepackage}[2][]{%
  \@namedef{ver@#2.sty}{9999/12/31}%
  \@namedef{opt@#2.sty}{#1}%
}
\makeatother

\dontusepackage{algorithm} 

\usepackage{microtype}
\usepackage{graphicx}
\usepackage{subcaption}
\usepackage{booktabs} 
\usepackage{kotex}
\usepackage{hyperref}
\usepackage[table]{xcolor}
\usepackage{bm}
\usepackage{xcolor}
\usepackage{multirow}

\usepackage[ruled,vlined]{algorithm2e}
\usepackage{array}

\usepackage[preprint]{icml2026}

\usepackage{amsmath}
\usepackage{amssymb}
\usepackage{mathtools}
\usepackage{amsthm}
\usepackage{listings}
\usepackage[dvipsnames]{xcolor} 

\usepackage[capitalize,noabbrev]{cleveref}

\theoremstyle{plain}

\theoremstyle{definition}

\theoremstyle{remark}

\usepackage{enumitem}
\usepackage[textsize=tiny]{todonotes}

\icmltitlerunning{STAG: Structural Test-time Alignment of Gradients for Online Adaptation}

\begin{document}

\twocolumn[
  \icmltitle{STAG: Structural Test-time Alignment of Gradients for Online Adaptation}




  \icmlsetsymbol{equal}{*}

    \begin{icmlauthorlist}
      \icmlauthor{Juhyeon Shin}{snu}
      \icmlauthor{Yujin Oh}{ewha}
      \icmlauthor{Jonghyun Lee}{krafton}
      \icmlauthor{Saehyung Lee}{ms}
      \icmlauthor{Minjun Park}{snu}
      \icmlauthor{Dongjun Lee}{gatech}
      \icmlauthor{Uiwon Hwang}{ewha}
      \icmlauthor{Sungroh Yoon}{snu}
    \end{icmlauthorlist}
    
    \icmlaffiliation{snu}{Seoul National University, Seoul, Republic of Korea}
    \icmlaffiliation{ewha}{Division of Artificial Intelligence \& Software, Ewha Womans University, Seoul, Republic of Korea}
    \icmlaffiliation{krafton}{KRAFTON, Seoul, Republic of Korea}
    \icmlaffiliation{ms}{Microsoft, Redmond, WA, USA}
    \icmlaffiliation{gatech}{Georgia Institute of Technology, Atlanta, GA, USA}
    
    \icmlcorrespondingauthor{Uiwon Hwang}{uiwon.hwang@ewha.ac.kr}
    \icmlcorrespondingauthor{Sungroh Yoon}{sryoon@snu.ac.kr}

  \icmlkeywords{Machine Learning, ICML}

  \vskip 0.3in
]



\printAffiliationsAndNotice{}  

\begin{abstract}
Test-Time Adaptation (TTA) adapts pre-trained models using only unlabeled test streams, requiring real-time inference and update without access to source data. We propose \textbf{Structural Test-time Alignment of Gradients (STAG)}, a lightweight \textbf{plug-in enhancer} that exploits an always-available structural signal: the classifier’s intrinsic geometry. STAG derives class-wise \emph{structural anchors} from classifier weights via self-structural entropy, and during adaptation analytically computes the predicted-class entropy gradient from forward-pass quantities, aligning it to the corresponding anchor with a cosine-similarity loss. This closed-form design incurs near-zero memory and latency overhead and requires no additional backpropagation beyond the underlying baseline. Across corrupted image classification and continual semantic segmentation, STAG provides broadly applicable performance gains for strong TTA baselines on both CNN and Transformer architectures regardless of the underlying normalization scheme, with particularly large gains under challenging online regimes such as imbalanced label shifts, single-sample adaptation, mixed corruption streams and long-horizon continual TTA.
\end{abstract}

\section{Introduction}
\label{sec:intro}
Deep networks are typically trained under the assumption that train and test distributions match~\cite{krizhevsky2017imagenet}, yet real-world corruptions (e.g., adverse weather and natural distortions) often violate this assumption and can severely degrade performance~\cite{hendrycks2019benchmarking}. While prior adaptation/generalization methods have been studied extensively~\cite{csurka2017domain,muandet2013domain,ganin2015unsupervised,shot,lee2022confidence,hwang2024sf}, many require source data access or costly offline adaptation stages. \\
Test-Time Adaptation (TTA) has emerged as a more practical paradigm, adapting models on-the-fly using only unlabeled test streams.
Unlike standard domain adaptation, TTA operates under extreme information scarcity: target labels are unavailable, source data are typically inaccessible, and the target distribution may evolve over time. Moreover, adaptation must be performed online, interleaved with inference, since predictions are required continuously while the model is being updated on the same incoming stream. These constraints make test-time learning fundamentally different from offline finetuning and place a premium on methods that are lightweight, stable, and compatible with real-world deployment budgets. \\
In this work, we focus on a source of signal that remains always available in information-scarce TTA: the model’s classifier geometry. Even when target labels and source data are absent, the classifier weights still encode class-specific structure learned during source training. Concretely, if we treat each class weight vector as a virtual input and pass it through the classifier, we observe a strikingly consistent pattern of structural correctness—each weight strongly matches its own class—across diverse architectures and normalization schemes. This suggests that, while test-time predictions and confidence may fluctuate under distribution shift, the classifier can still provide a stable structural prior. We formalize this intuition via self-structural entropy and use it to derive class-wise reference directions that can guide online updates without requiring any additional supervision. \\
Based on this, we propose \textbf{Structural Test-time Alignment of Gradients (STAG)}, a plug-and-play regularizer applicable to arbitrary TTA objectives. STAG (i) extracts class-specific structural anchors from classifier weights in closed form before adaptation, and (ii) during online adaptation, aligns the predicted-class entropy gradient with the corresponding anchor via a cosine-similarity loss, weighted by a decaying coefficient. Both components admit efficient analytical forms, adding negligible overhead and requiring no additional backpropagation beyond the underlying baseline. \\
We evaluate STAG on image classification (ImageNet-C, CIFAR-10/100-C) and semantic segmentation (Cityscapes-ACDC) across CNNs and Transformers with diverse normalization schemes. STAG delivers robust performance enhancements under both stationary and dynamic shifts, including single-sample adaptation, label imbalance, mixed corruption streams and long-horizon continual TTA, while maintaining near-zero latency.
The primary contributions of this work are summarized as follows:
\begin{itemize}
    \item \textbf{Plug-and-play Regularization:} We propose \textbf{STAG}, a broadly applicable \textbf{plug-in enhancer} for online test-time adaptation. STAG exploits always-available \emph{classifier structure} to supply an additional, architecture-agnostic signal that can be seamlessly combined with existing TTA objectives.
    \item \textbf{Analytical Efficiency:} STAG admits a \textbf{closed-form} formulation for computing structural anchors and the associated alignment signal, incurring \textbf{near-zero} memory and latency overhead. This makes it practical for real-time online deployment without introducing additional backpropagation beyond the underlying baseline.
    \item \textbf{Architecture-Agnostic Robustness:} We validate STAG across both CNNs and Transformers with diverse normalization schemes (BN, GN, LN), and demonstrate that STAG yields significant performance improvements across a broad spectrum of scenarios, encompassing imbalanced label shifts, single-sample adaptation, mixed corruption settings and long-horizon continual TTA.
\end{itemize}

\section{Related Work}
\label{sec:related}
Test-Time Adaptation (TTA) updates a deployed model on-the-fly using only an unlabeled test stream, while maintaining tight latency and memory budgets. 
To satisfy these constraints, many methods adapt only lightweight components such as normalization statistics or affine parameters at test time~\cite{schneider2020improving}. 
A widely used learning signal is \emph{entropy minimization}, exemplified by TENT~\cite{wang2020tent} and its efficient variant EATA~\cite{eata}. 
More recent baselines such as SAR~\cite{sar} and DeYO~\cite{deyo} further improve robustness by incorporating additional regularization and/or sample filtering mechanisms on top of entropy-based adaptation. 
Beyond the standard setting, \emph{continual} TTA considers long, non-stationary test streams where the distribution changes over time and naïve online updates can accumulate errors and cause forgetting; methods such as CoTTA~\cite{wang2022cotta} and EcoTTA~\cite{song2023ecotta} explicitly target stability over extended horizons by introducing temporal ensembling and/or consistency-based regularization across test-time states. 
In contrast to proposing another standalone objective, we present \textbf{STAG} as a lightweight \emph{plug-in enhancer} that can be seamlessly combined with these TTA baselines by exploiting an always-available structural signal from the classifier geometry to guide online updates.

\section{Structural Test-time Alignment of Gradients}
\begin{figure*}[t]
    \centering
    \includegraphics[width=0.9\textwidth]{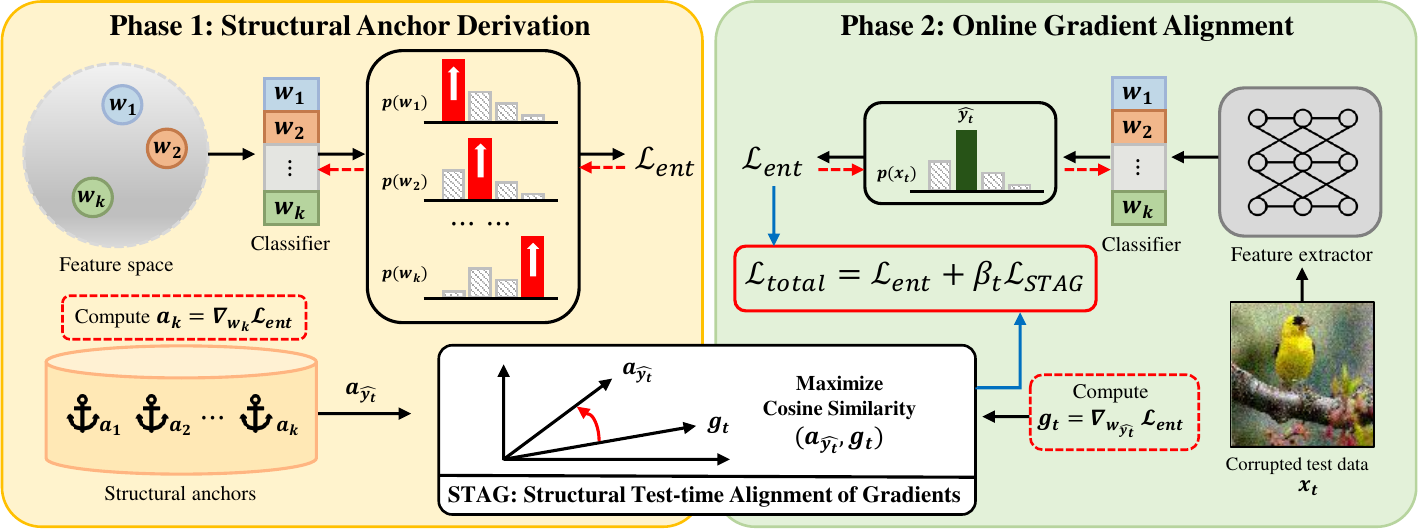}
    \caption{\textbf{The Overall Framework of STAG.} 
    (a) \textbf{Phase 1: Structural Anchor Derivation.} We derive structural anchors $\mathbf{a}_k$ by calculating the gradient of self-structural entropy from the classifier weights. 
    (b) \textbf{Phase 2: Online Gradient Alignment.} During test-time adaptation, STAG analytically computes the online entropy gradient $\mathbf{g}_t$ from test samples and aligns it with the corresponding structural anchor $\mathbf{a}_{\hat{y}_t}$.}
    \label{fig:overview}
\end{figure*}
\subsection{Preliminaries}
\label{sec:prelim}
\paragraph{Problem Setup} 
We consider the task of \textbf{Test-Time Adaptation (TTA)} , where a model pre-trained on a source domain is adapted to an unseen target domain without any access to the original source data. Let the source training and target test datasets be denoted as $\mathcal{D}_{\mathrm{tr}} = \{(\mathbf{x}_i^{\mathrm{tr}}, y_i^{\mathrm{tr}}): \mathbf{x}_i^{\mathrm{tr}} \sim P(\mathcal{X}), y_i^{\mathrm{tr}} \in \mathcal{Y}\}$ and $\mathcal{D}_{\mathrm{te}} = \{(\mathbf{x}_i^{\mathrm{te}}, y_i^{\mathrm{te}}): \mathbf{x}_i^{\mathrm{te}} \sim Q(\mathcal{X}), y_i^{\mathrm{te}} \in \mathcal{Y}\}$, respectively. In the TTA setting, we encounter a distribution shift $P(\mathcal{X}) \neq Q(\mathcal{X})$ while the target labels $y_i^{\mathrm{te}}$ remain strictly unavailable during the online adaptation process. Both domains are assumed to share a common label space $\mathcal{Y} = \{1, \dots, c\}$ consisting of $c$ distinct classes. This setup poses a fundamental challenge: the model must effectively mitigate the domain gap by maximizing the utility of its internal knowledge under extreme information scarcity.
\vspace{-2mm}
\paragraph{Model Architecture}
The model consists of a feature extractor $f_{\bm{\phi}}: \mathcal{X} \rightarrow \mathcal{H}$ with parameters $\bm{\phi}$, which maps input data to a $D$-dimensional feature space $\mathcal{H} \subseteq \mathbb{R}^D$. This is followed by a classifier $g_{\mathbf{w}}: \mathcal{H} \rightarrow \mathbb{R}^c$ parameterized by $\mathbf{w} = \{\mathbf{b}, \mathbf{w}_1, \dots, \mathbf{w}_c\}$. Each weight vector $\mathbf{w}_k \in \mathbb{R}^D$ corresponds to class $k \in \{1, \dots, c\}$ and encapsulates the semantic representation of that category. For a given input $\mathbf{x}$, the model generates a feature representation $\mathbf{h} = f_{\bm{\phi}}(\mathbf{x}) \in \mathcal{H}$ and a subsequent logit vector $\mathbf{z} = g_{\mathbf{w}}(\mathbf{h}) \in \mathbb{R}^c$. The final prediction probability $\mathbf{p} = \sigma(\mathbf{z})$ is obtained via the softmax function $\sigma(\cdot)$, where the $k$-th element is defined as $p_k = \exp(z_k) / \sum_{j=1}^c \exp(z_j)$.
\vspace{-2mm}
\paragraph{Online Test-Time Adaptation}
During the online adaptation phase, test samples $\{\mathbf{x}_t^{\mathrm{te}}\}$ arrive sequentially, and the model updates a subset of its parameters $\theta_{\mathrm{adapt}} \subseteq \{\bm{\phi}, \mathbf{w}\}$. Due to the absence of ground-truth labels in the target domain, it is standard practice~\cite{wang2020tent, eata, sar, deyo} to minimize an unsupervised objective such as the Shannon entropy loss $\mathcal{L}_{\mathrm{ent}}$:
\begin{equation}
\mathcal{L}_{\mathrm{ent}}(\mathbf{x}) = -\sum_{i=1}^{c} p_i \ln p_i.
\label{eq:ent}\end{equation}
Rather than relying solely on the instantaneous entropy gradient, we propose a geometric regularizer that aligns the adaptation direction with structural anchors derived from the classifier's own weights. By providing a reference based on the classifier's intrinsic geometry, this alignment mechanism maximizes the utility of internal information and facilitates the formation of a feature space that remains semantically compatible with the classifier’s decision logic.
\vspace{-2mm}
\subsection{Methodology}
\label{sec:method}
In this section, we present the details of \textbf{Structural Test-time Alignment of Gradients (STAG)}. STAG implements a two-phase process: (i) extracting structural anchors derived from the classifier weights , and (ii) aligning the online gradients with these anchors to guide the adaptation. The overall framework is illustrated in Figure~\ref{fig:overview}.
\vspace{-2mm}
\paragraph{Anchor Extraction from Self-structural Entropy}
\begin{figure}[t]
    \centering
    \includegraphics[width=1.0\columnwidth]{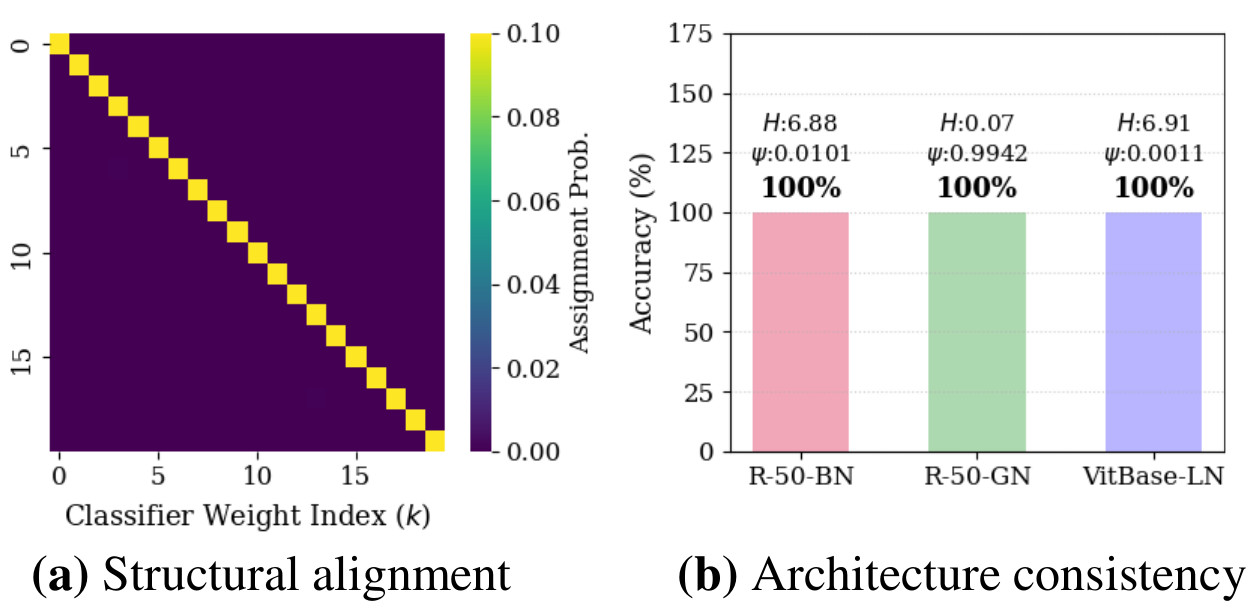}
    \vspace{-3mm} 
\caption{\textbf{Validation of classifier weights as structural anchors.} (a) Softmax assignment probability heatmap for 20 selected ImageNet classes (ResNet50-GN). (b) Metric consistency across architectures: ResNet50-BN, ResNet50-GN and VitBase-LN. $H$, $\psi$, and Acc denote self-entropy, mean diagonal probability (MDP), and structural accuracy, respectively.}
    \label{fig:structural_anchor}
    \vspace{-5mm}
\end{figure}
STAG treats the classifier weights $\{\mathbf{w}_1, \dots, \mathbf{w}_c\}$ as meaningful proxies for the semantic structure of the feature space. To justify this, we analyze the structural consistency by treating each weight vector $\mathbf{w}_k$ as a virtual input to the classifier. As shown in Figure~\ref{fig:structural_anchor}(a), the softmax heatmap for 20 sampled classes from a ResNet50-GN pretrained on ImageNet exhibits a sharp diagonal, confirming that each weight is correctly mapped to its respective category. Furthermore, our evaluation across various architectures—including ResNet with Batch Normalization (BN), Group Normalization (GN), and ViT with Layer Normalization (LN)—reveals that even when the mean diagonal probability ($\psi$) is low or self-entropy ($H$) is high, the \textit{structural accuracy} remains consistently at 100\% (see Figure~\ref{fig:structural_anchor}(b)). This 100\% accuracy demonstrates that the classifier weights, despite differing in their probabilistic confidence, preserve the essential semantic boundaries of the inherent classifier geometry.

Based on this insight, we extract the structural relationships by defining an \textbf{inter-class similarity matrix} $\mathbf{Z} \in \mathbb{R}^{c \times c}$. Each element $Z_{kj}$ represents the logit value for the $j$-th class when the $k$-th class weight $\mathbf{w}_k$ is used as a virtual input to the classifier:
\begin{equation}
Z_{kj} = \mathbf{w}_j^\top \mathbf{w}_k + b_j,
\label{eq:inter-class_matrix}
\end{equation}
where $b_j$ is the bias term corresponding to the $j$-th class. Then we calculate the \textbf{self-structural entropy} $H_k$:
\begin{equation}
H_k = -\sum_{j=1}^c p_{kj} \ln p_{kj},
\label{eq:self_entropy}
\end{equation}
where $p_{kj} = \sigma(\mathbf{Z}_k)_j$ is the softmax probability.

The structural anchor $\mathbf{a}_k$ is then derived by isolating the \textbf{self-structural sensitivity}, which represents the most dominant directional component for minimizing structural ambiguity. While the full gradient of entropy involves a linear combination of all weights, we focus on the sensitivity of the $k$-th class with respect to its own prototype to ensure a de-noised and class-specific guidance.
Specifically, $\mathbf{a}_k$ is formulated in an analytical closed-form as follows:
\begin{equation}
\mathbf{a}_k = s_k \mathbf{w}_k, \quad \text{where} \quad s_k = -p_{kk} (\ln p_{kk} + H_k).
\label{eq:structural_anchor}
\end{equation}
Here, the \textbf{structural sensitivity $s_k$} captures the internal gradient magnitude that stabilizes the $k$-th decision boundary. Detailed derivation of this component is provided in Appendix \ref{sec:appendix_derivation}. This self-referential formulation ensures both computational efficiency and stable adaptation by filtering out gradient interference from unrelated classes.

\vspace{-2mm}
\paragraph{Sample-wise Gradient Extraction}
During the online adaptation stage, STAG identifies the instantaneous update direction for each arriving test sample $\mathbf{x}_t$. Specifically, we first obtain the feature representation $\mathbf{h}_t = f_{\bm{\phi}}(\mathbf{x}_t) \in \mathbb{R}^D$ and the predicted class $\hat{y}_t = \arg\max_k p_{tk}$ through a standard forward pass. To determine the model's adaptation intent, we extract the gradient of the entropy loss $\mathcal{L}_{\text{ent}}$ with respect to the weight vector of the predicted class, $\mathbf{w}_{\hat{y}_t}$.
Following the analytical derivation in Appendix \ref{sec:appendix_derivation}, the \textbf{logit-level sensitivity} $s_t$ for the sample is defined as:
\begin{equation}
s_t = \frac{\partial \mathcal{L}_{\text{ent}}}{\partial z_{t,\hat{y}_t}} = -p_{t,\hat{y}_t} (\ln p_{t,\hat{y}_t} + H(\mathbf{p}_t)),
\label{eq:sample_sensitivity}
\end{equation}
where $H(\mathbf{p}_t)$ is the predictive entropy of the current sample. The resulting \textbf{online gradient} $\mathbf{g}_t$, which represents the sample-driven adaptation direction in the weight geometry, is then formulated as:
\begin{equation}
\mathbf{g}_t = \nabla_{\mathbf{w}_{\hat{y}_t}} \mathcal{L}_{\text{ent}} = s_t \cdot \mathbf{h}_t.
\label{eq:online_gradient}
\end{equation}
This closed-form formulation allows $\mathbf{g}_t$ to be computed directly from the forward-pass quantities ($p_t$ and $\mathbf{h}_t$) without any additional backpropagation. Consequently, STAG captures fine-grained adaptation behavior at each time step with negligible computational overhead, providing the vector required for subsequent structural alignment.

\paragraph{Structural Gradient Alignment}
The final objective of STAG is to rectify the instantaneous adaptation direction by aligning it with the model's intrinsic structural knowledge. Given the online gradient $\mathbf{g}_t$ and the corresponding structural anchor $\mathbf{a}_{\hat{y}_t}$, we define the alignment loss $\mathcal{L}_{\text{STAG}}$ as the negative cosine similarity between the two vectors:
\begin{equation}
\mathcal{L}_{\text{STAG}}(\mathbf{x}_t) = -\frac{\mathbf{g}_t \cdot \mathbf{a}_{\hat{y}_t}}{\|\mathbf{g}_t\| \|\mathbf{a}_{\hat{y}_t}\| + \epsilon},
\label{eq:stag_loss}
\end{equation}
where $\epsilon$ is a small constant for numerical stability. 

For a mini-batch $\mathcal{B}_t$ arriving at step $t$, the batch-wise alignment loss is the empirical mean over the samples:
\begin{equation}
\mathcal{L}_{\text{STAG}}(\mathcal{B}_t) = \frac{1}{|\mathcal{B}_t|} \sum_{\mathbf{x} \in \mathcal{B}_t} \mathcal{L}_{\text{STAG}}(\mathbf{x}).
\label{eq:stag_batch_loss}
\end{equation}
The final objective to be minimized at step $t$ is formulated as a weighted combination of a baseline TTA loss and the proposed $\mathcal{L}_{\text{STAG}}$:
\begin{equation}
    \mathcal{L}_{\text{total}} = \mathcal{L}_{\text{base}}(\mathcal{B}_t) + \beta_t \mathcal{L}_{\text{STAG}}(\mathcal{B}_t),
    \label{eq:total_objective}
\end{equation}
where $\mathcal{L}_{\text{base}}$ denotes an arbitrary unsupervised objective used in standard TTA frameworks (e.g., Shannon entropy in TENT~\cite{wang2020tent} or importance-weighted entropy in EATA~\cite{eata}). Here, $\beta_t$ is a dynamic balancing coefficient that follows an exponential decay:
\begin{equation}
    \beta_{t+1} = \beta_t \cdot \exp(-1/\gamma).
    \label{eq:adaptive_decay}
\end{equation}
This decaying strategy ensures strong structural guidance in the early stages of adaptation to prevent initial semantic drift, while gradually relaxing the constraint as the model becomes more familiar with the target domain. By framing STAG as a structural regularizer, it can be seamlessly integrated into various baseline TTA methods to enhance their architectural consistency and adaptation stability. The entire procedure is summarized in Appendix~\ref{sec:append_add} (Algorithm~\ref{alg:stag}).
\subsection{Computational Complexity}
\label{sec:complexity}
Compared to standard TTA methods, the computational overhead of STAG is virtually negligible due to its analytical nature. The complexity varies slightly depending on whether the classifier weights $W$ are updated:
\vspace{-2mm}
\begin{itemize}
    \item \textbf{Static Classifier:} If the classifier $W$ remains frozen (e.g., when only \textit{affine parameters} of the normalization layers are adapted), the structural anchors are computed only \textit{once} offline. The online overhead is then strictly limited to vector indexing and cosine similarity calculations, adding a negligible $O(|\mathcal{B}_t| \cdot D)$ per batch.
    \item \textbf{Dynamic Classifier:} If $W$ is updated during adaptation, the anchors $\mathbf{a}_k$ are recalculated to maintain geometric consistency. However, since the sensitivity $s_k$ is derived from a closed-form analytical solution (Eq.~\ref{eq:structural_anchor}), this process requires only a single matrix multiplication $WW^\top$ and element-wise operations. 
\end{itemize}

Crucially, this recalculation is equivalent to a single "self-forward pass" of the classifier and \textbf{does not increase the number of backward passes}. Since the cost of a full backward pass through the network parameters $\theta$ is significantly higher than a classifier-level matrix multiplication ($|\theta| \gg C^2 D$), STAG provides structural guidance with near-zero latency compared to the baseline adaptation time.

\vspace{-2mm}
\section{Experiments}
In this section, we evaluate the efficacy of STAG across a diverse range of test-time adaptation scenarios, including image classification and semantic segmentation tasks, while further extending our evaluation to challenging regimes such as imbalanced label shifts, single-sample adaptation, mixed corruption streams, and long-horizon continual TTA.
\definecolor{LightPink}{RGB}{255,238,238}

\begin{table*}[t]
\centering
\small
\setlength{\tabcolsep}{3pt}
\renewcommand{\arraystretch}{1.15}
\caption{\textbf{Classification accuracy (\%) on ImageNet-C (severity 5) under the mild scenario.} For each baseline and its STAG-integrated version (+STAG), the superior performance is highlighted in \textbf{bold}. Values in parentheses indicate the absolute accuracy gain achieved by STAG.}
\label{tab:mild}
\resizebox{\textwidth}{!}{%
\begin{tabular}{l|ccc|cccc|cccc|cccc|c}
\multicolumn{1}{l}{} & \multicolumn{3}{c}{Noise} & \multicolumn{4}{c}{Blur} & \multicolumn{4}{c}{Weather} & \multicolumn{4}{c}{Digital} & \\
\textbf{Mild} & Gauss. & Shot & Impul. & Defoc. & Glass & Motion & Zoom & Snow & Frost & Fog & Brit. & Contr. & Elastic & Pixel & JPEG & Avg. \\
\midrule
ResNet50-BN & 2.2 & 2.9 & 1.8 & 17.9 & 9.8 & 14.8 & 22.5 & 16.9 & 23.3 & 24.4 & 58.9 & 5.4 & 16.9 & 20.6 & 31.7 & 18.0 \\
\textbullet\ TENT & 28.6 & 30.5 & 30.0 & 28.0 & 27.3 & 41.2 & 49.2 & 47.1 & 41.1 & 57.6 & \textbf{67.4} & 27.0 & 54.6 & 58.5 & 52.3 & 42.7 \\
\rowcolor{LightPink} +STAG & $\textbf{30.5}_{\pm0.2}$ & $\textbf{32.3}_{\pm0.5}$ & $\textbf{32.3}_{\pm0.2}$ & $\textbf{30.1}_{\pm0.2}$ & $\textbf{29.2}_{\pm0.5}$ & $\textbf{44.6}_{\pm0.2}$ & $\textbf{50.7}_{\pm0.2}$ & $\textbf{49.6}_{\pm0.2}$ & $\textbf{43.3}_{\pm0.0}$ & $\textbf{58.6}_{\pm0.2}$ & $67.3_{\pm0.1}$ & $\textbf{31.4}_{\pm0.5}$ & $\textbf{56.4}_{\pm0.2}$ & $\textbf{59.7}_{\pm0.1}$ & $\textbf{53.8}_{\pm0.2}$ & $\textbf{44.7}_{\pm0.1}$ {\scriptsize(+1.96)} \\
\textbullet\ EATA & 34.9 & 36.9 & 35.7 & 33.5 & 33.1 & 47.0 & 52.7 & 51.3 & 45.5 & 59.9 & \textbf{67.9} & 44.6 & 57.9 & 60.4 & 54.9 & 47.7 \\
\rowcolor{LightPink} +STAG & $\textbf{35.6}_{\pm0.1}$ & $\textbf{38.0}_{\pm0.2}$ & $\textbf{36.7}_{\pm0.3}$ & $\textbf{34.6}_{\pm0.3}$ & $\textbf{34.1}_{\pm0.3}$ & $\textbf{48.9}_{\pm0.1}$ & $\textbf{53.2}_{\pm0.1}$ & $\textbf{52.7}_{\pm0.1}$ & $\textbf{46.5}_{\pm0.2}$ & $\textbf{60.4}_{\pm0.1}$ & $67.6_{\pm0.0}$ & $\textbf{46.4}_{\pm0.1}$ & $\textbf{58.5}_{\pm0.2}$ & $\textbf{60.8}_{\pm0.0}$ & $\textbf{55.5}_{\pm0.2}$ & $\textbf{48.6}_{\pm0.0}$ {\scriptsize(+0.89)} \\
\textbullet\ SAR & 30.7 & 30.4 & 31.1 & 28.9 & 28.9 & 41.8 & 49.3 & 47.2 & 42.2 & 57.5 & 67.3 & 37.2 & 54.5 & 58.4 & 52.3 & 43.8 \\
\rowcolor{LightPink} +STAG & $\textbf{32.9}_{\pm0.2}$ & $\textbf{33.7}_{\pm0.2}$ & $\textbf{33.4}_{\pm0.5}$ & $\textbf{31.3}_{\pm0.5}$ & $\textbf{31.4}_{\pm0.4}$ & $\textbf{45.3}_{\pm0.1}$ & $\textbf{51.1}_{\pm0.1}$ & $\textbf{50.2}_{\pm0.1}$ & $\textbf{44.1}_{\pm0.2}$ & $\textbf{59.0}_{\pm0.1}$ & $\textbf{67.7}_{\pm0.1}$ & $\textbf{40.8}_{\pm1.1}$ & $\textbf{56.7}_{\pm0.1}$ & $\textbf{60.1}_{\pm0.1}$ & $\textbf{54.2}_{\pm0.2}$ & $\textbf{46.1}_{\pm0.1}$ {\scriptsize(+2.29)} \\
\textbullet\ DeYO & 35.6 & 38.1 & 36.8 & 33.8 & 33.8 & 48.4 & 52.8 & 52.6 & 46.2 & 60.6 & \textbf{68.0} & \textbf{44.6} & 58.4 & 61.4 & 55.7 & 48.5 \\
\rowcolor{LightPink} +STAG & $\textbf{35.9}_{\pm0.1}$ & $\textbf{38.3}_{\pm0.1}$ & $\textbf{37.0}_{\pm0.4}$ & $\textbf{34.1}_{\pm0.4}$ & $\textbf{34.1}_{\pm0.4}$ & $\textbf{49.0}_{\pm0.1}$ & $\textbf{53.0}_{\pm0.0}$ & $\textbf{53.1}_{\pm0.1}$ & $\textbf{46.5}_{\pm0.3}$ & $\textbf{60.8}_{\pm0.1}$ & $67.8_{\pm0.1}$ & $44.0_{\pm3.9}$ & $\textbf{58.6}_{\pm0.0}$ & $\textbf{61.5}_{\pm0.1}$ & $\textbf{55.8}_{\pm0.1}$ & $\textbf{48.6}_{\pm0.3}$ {\scriptsize(+0.17)} \\
\midrule
ResNet50-GN & 18.0 & 19.8 & 17.9 & 19.8 & 11.3 & 21.4 & 24.9 & 40.4 & 47.3 & 33.6 & 69.3 & 36.3 & 18.6 & 28.4 & 52.3 & 30.6 \\
\textbullet\ TENT & 5.0 & 6.6 & 5.7 & 15.1 & 7.9 & 21.9 & 22.7 & 26.3 & 33.2 & 3.3 & 70.3 & 42.4 & 11.0 & 48.1 & 54.3 & 24.9 \\
\textbullet\ EATA & 37.5 & 39.9 & 38.8 & 28.3 & 27.2 & 36.7 & 39.1 & 50.9 & 49.4 & 55.2 & 72.0 & 50.1 & 41.8 & 55.7 & 58.1 & 45.4 \\
\rowcolor{LightPink} +STAG & $\textbf{44.1}_{\pm0.3}$ & $\textbf{46.1}_{\pm0.3}$ & $\textbf{44.9}_{\pm0.3}$ & $\textbf{34.0}_{\pm1.4}$ & $\textbf{35.6}_{\pm0.6}$ & $\textbf{46.0}_{\pm0.3}$ & $\textbf{49.5}_{\pm0.2}$ & $\textbf{58.8}_{\pm0.2}$ & $\textbf{54.8}_{\pm0.2}$ & $\textbf{63.3}_{\pm0.5}$ & $\textbf{73.2}_{\pm0.0}$ & $\textbf{55.3}_{\pm0.1}$ & $\textbf{55.5}_{\pm0.4}$ & $\textbf{63.1}_{\pm0.1}$ & $\textbf{60.9}_{\pm0.1}$ & $\textbf{52.3}_{\pm0.1}$ {\scriptsize(+6.97)} \\
\textbullet\ SAR & 28.2 & 31.3 & 29.6 & \textbf{18.8} & 19.3 & 30.4 & 30.5 & 41.8 & 43.3 & 44.4 & 70.7 & 43.9 & \textbf{17.1} & 48.7 & 55.2 & 36.9 \\
\rowcolor{LightPink} +STAG & $\textbf{39.1}_{\pm0.0}$ & $\textbf{42.3}_{\pm0.2}$ & $\textbf{40.8}_{\pm0.3}$ & $17.5_{\pm0.3}$ & $\textbf{23.5}_{\pm1.0}$ & $\textbf{36.6}_{\pm0.6}$ & $\textbf{39.5}_{\pm0.7}$ & $\textbf{51.2}_{\pm0.4}$ & $\textbf{48.8}_{\pm0.2}$ & $\textbf{52.8}_{\pm3.0}$ & $\textbf{73.2}_{\pm0.1}$ & $\textbf{49.3}_{\pm0.1}$ & $1.4_{\pm0.2}$ & $\textbf{55.8}_{\pm0.2}$ & $\textbf{56.7}_{\pm0.1}$ & $\textbf{41.9}_{\pm0.3}$ {\scriptsize(+5.01)} \\
\textbullet\ DeYO & 39.6 & 42.3 & 40.9 & 22.7 & 24.7 & 38.9 & 37.2 & 51.6 & 49.8 & 38.3 & 73.2 & 50.4 & 42.9 & 56.6 & 58.0 & 44.5 \\
\rowcolor{LightPink} +STAG & $\textbf{41.4}_{\pm0.2}$ & $\textbf{44.2}_{\pm0.4}$ & $\textbf{42.7}_{\pm0.5}$ & $\textbf{22.9}_{\pm0.1}$ & $\textbf{25.5}_{\pm0.3}$ & $\textbf{40.1}_{\pm0.1}$ & $\textbf{37.8}_{\pm1.4}$ & $\textbf{53.6}_{\pm0.2}$ & $\textbf{51.3}_{\pm0.1}$ & $\textbf{57.8}_{\pm1.0}$ & $\textbf{73.5}_{\pm0.1}$ & $\textbf{51.4}_{\pm0.2}$ & $\textbf{43.0}_{\pm1.6}$ & $\textbf{57.9}_{\pm0.2}$ & $\textbf{58.4}_{\pm0.1}$ & $\textbf{46.8}_{\pm0.1}$ {\scriptsize(+2.29)} \\
\midrule
VitBase-LN & 9.5 & 6.7 & 8.2 & 29.0 & 23.4 & 33.9 & 27.1 & 15.9 & 26.5 & 47.2 & 54.7 & 44.1 & 30.5 & 44.5 & 47.8 & 29.9 \\
\textbullet\ TENT & 42.1 & \textbf{1.3} & 43.0 & 52.3 & 47.7 & 55.2 & 49.9 & 18.7 & 21.6 & 66.3 & 74.7 & 64.7 & 51.8 & 66.6 & 63.9 & 48.0 \\
\rowcolor{LightPink} +STAG & $\textbf{52.1}_{\pm0.0}$ & $0.3_{\pm0.1}$ & $\textbf{53.1}_{\pm0.1}$ & $\textbf{57.6}_{\pm0.1}$ & $\textbf{57.9}_{\pm0.2}$ & $\textbf{62.3}_{\pm0.1}$ & $\textbf{59.8}_{\pm0.2}$ & $\textbf{65.3}_{\pm1.1}$ & $\textbf{52.4}_{\pm15.1}$ & $\textbf{73.0}_{\pm0.1}$ & $\textbf{77.5}_{\pm0.0}$ & $\textbf{68.0}_{\pm0.2}$ & $\textbf{66.7}_{\pm0.1}$ & $\textbf{72.5}_{\pm0.1}$ & $\textbf{69.6}_{\pm0.1}$ & $\textbf{59.2}_{\pm1.1}$ {\scriptsize(+11.21)} \\
\textbullet\ EATA & 50.4 & 50.0 & 51.2 & 55.7 & 55.8 & 60.0 & 57.6 & 63.6 & 61.9 & 71.3 & 76.1 & 66.9 & 64.5 & 70.4 & 67.5 & 61.5 \\
\rowcolor{LightPink} +STAG & $\textbf{54.9}_{\pm0.3}$ & $\textbf{55.8}_{\pm0.3}$ & $\textbf{55.7}_{\pm0.3}$ & $\textbf{58.4}_{\pm0.2}$ & $\textbf{59.6}_{\pm0.1}$ & $\textbf{64.6}_{\pm0.3}$ & $\textbf{63.6}_{\pm0.2}$ & $\textbf{68.8}_{\pm0.3}$ & $\textbf{67.1}_{\pm0.3}$ & $\textbf{73.7}_{\pm0.2}$ & $\textbf{77.4}_{\pm0.2}$ & $\textbf{68.2}_{\pm0.1}$ & $\textbf{69.7}_{\pm0.2}$ & $\textbf{73.1}_{\pm0.3}$ & $\textbf{70.0}_{\pm0.2}$ & $\textbf{65.4}_{\pm0.2}$ {\scriptsize(+3.85)} \\
\textbullet\ SAR & 43.8 & 33.7 & 44.8 & 52.9 & 50.0 & 55.9 & 51.2 & 57.3 & 55.6 & 66.3 & 74.7 & 64.4 & 55.1 & 66.6 & 63.9 & 55.8 \\
\rowcolor{LightPink} +STAG & $\textbf{54.0}_{\pm0.2}$ & $\textbf{53.3}_{\pm2.2}$ & $\textbf{53.1}_{\pm3.3}$ & $\textbf{58.2}_{\pm0.1}$ & $\textbf{59.3}_{\pm0.2}$ & $\textbf{63.5}_{\pm0.1}$ & $\textbf{62.0}_{\pm0.1}$ & $\textbf{67.5}_{\pm0.2}$ & $\textbf{66.0}_{\pm0.1}$ & $\textbf{73.6}_{\pm0.1}$ & $\textbf{77.8}_{\pm0.1}$ & $\textbf{67.8}_{\pm0.4}$ & $\textbf{68.7}_{\pm0.2}$ & $\textbf{73.2}_{\pm0.1}$ & $\textbf{70.2}_{\pm0.0}$ & $\textbf{64.5}_{\pm0.3}$ {\scriptsize(+8.79)} \\
\textbullet\ DeYO & 53.3 & 52.1 & 54.2 & 58.3 & 58.7 & 63.1 & 51.6 & 67.3 & 66.1 & 73.2 & \textbf{78.1} & \textbf{68.0} & 68.0 & 73.2 & \textbf{70.1} & 63.7 \\
\rowcolor{LightPink} +STAG & $\textbf{56.1}_{\pm0.3}$ & $\textbf{57.4}_{\pm0.2}$ & $\textbf{56.5}_{\pm0.5}$ & $\textbf{58.9}_{\pm0.2}$ & $\textbf{60.3}_{\pm0.1}$ & $\textbf{65.8}_{\pm0.0}$ & $\textbf{65.4}_{\pm0.1}$ & $\textbf{69.1}_{\pm0.5}$ & $\textbf{66.7}_{\pm0.6}$ & $\textbf{73.6}_{\pm0.6}$ & $77.6_{\pm0.1}$ & $67.2_{\pm0.3}$ & $\textbf{71.1}_{\pm0.0}$ & $\textbf{73.8}_{\pm0.2}$ & $70.0_{\pm0.6}$ & $\textbf{66.0}_{\pm0.1}$ {\scriptsize(+2.26)} \\
\end{tabular}}
\vspace{-2mm}
\end{table*}

\subsection{Experimental Setup}
\paragraph{Datasets} 
We evaluate STAG on two primary tasks to assess its versatility. For image classification, we use ImageNet-C and CIFAR-10/100-C~\cite{hendrycks2019benchmarking}, widely used benchmarks for evaluating model robustness and test-time adaptation. Both datasets consist of 15 corruption types with 5 severity levels. For semantic segmentation, we use the Cityscapes-to-ACDC benchmark, where Cityscapes~\cite{cordts2016cityscapes} is the source domain and the Adverse Conditions Dataset with Correspondences (ACDC)~\cite{sakaridis2021acdc} is the target domain with four weather corruptions: Fog, Night, Rain, and Snow.
\vspace{-2mm}
\paragraph{Architectures}
For CIFAR-10/100-C, we use WideResNet-28/40~\cite{zagoruyko2016wide}, respectively. For ImageNet-C, we evaluate ResNet50-BN, ResNet50-GN, and VitBase-LN, using models from \texttt{torchvision} or \texttt{timm}~\cite{rw2019timm}. For segmentation, we adopt a pretrained Segformer-B5~\cite{xie2021segformer}.
\vspace{-2mm}
\paragraph{Baselines}
We compare against representative TTA methods: \textbf{Source} (no adaptation), \textbf{TENT}~\cite{wang2020tent} (entropy minimization), \textbf{EATA}~\cite{eata} (stability regularization with re-weighting), \textbf{SAR}~\cite{sar} (sharpness-aware updates with noisy-sample filtering), and \textbf{DeYO}~\cite{deyo} (confidence-based filtering via PLPD). For continual TTA, we additionally consider \textbf{BN Adapt}~\cite{bn}, \textbf{CoTTA}~\cite{wang2022cotta}, and \textbf{EcoTTA}~\cite{song2023ecotta} (partition factor $\mathrm{K}\in\{4,5\}$).
\vspace{-2mm}
\paragraph{Implementation Details}
We follow each baseline's original optimization protocol for fair comparison. STAG uses two hyper-parameters, $\beta_0$ and $\gamma$, selected using only a single corruption type (the first corruption in our evaluation order) and then fix them for all remaining corruption types. The specific hyperparameter configurations for each experimental setting are detailed in Appendix ~\ref{sec:append_add} (Table ~\ref{tab:appendix_hp}).

\subsection{Main Results on Image Classification}
\begin{figure*}[t]
    \centering
    \vspace{-2mm} 
    \includegraphics[width=0.98\textwidth]{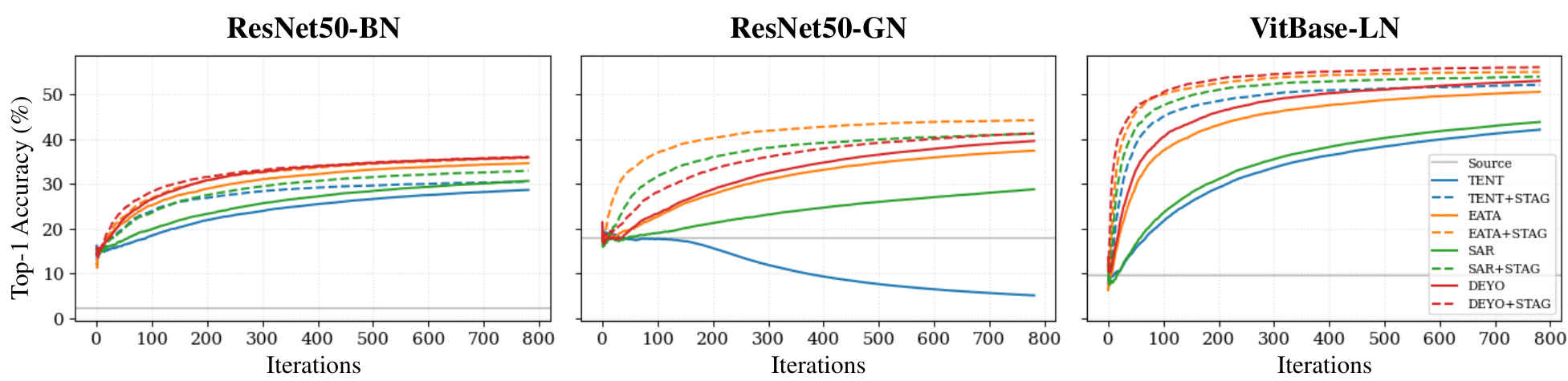}
    \vspace{-1mm} 
    \caption{\textbf{Top-1 accuracy running curves} across three different architectures on ImageNet-C Gaussian noise (severity 5)}
    \label{fig:running_curves}
    \vspace{-4mm} 
\end{figure*}
\begin{figure}[t]
  \centering
  \includegraphics[width=\columnwidth]{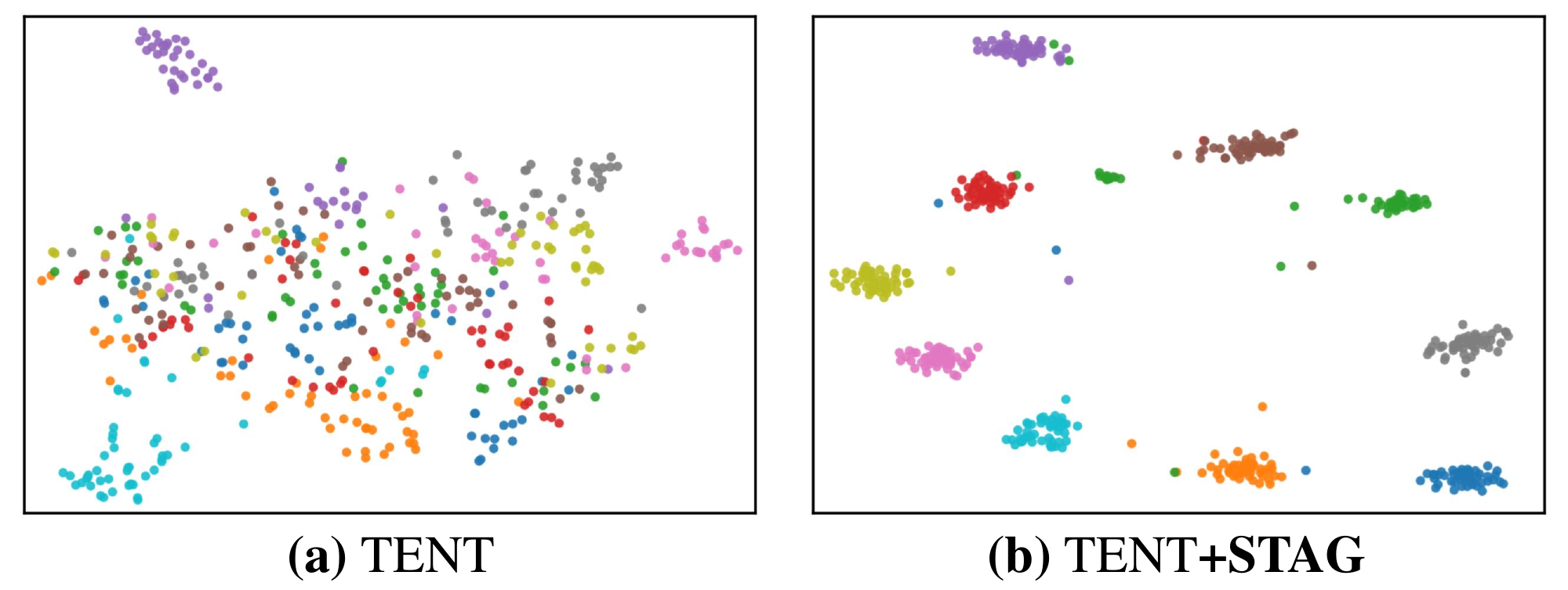}
  \caption{\textbf{The t-SNE visualization of the feature space} on ImageNet-C Frost (severity 5)}
  \label{fig:tsne}
\end{figure}
The \textbf{mild scenario} is a standard test-time adaptation setting with a large batch size ($B=64$) and a stationary distribution, serving as a benchmark before more extreme dynamic shifts. Table~\ref{tab:mild} summarizes the classification accuracy on ImageNet-C (severity 5). We observe that STAG improves all four baselines across architectures. In \textbf{ResNet50-BN}, STAG yields an average gain of up to \textbf{+2.29\%} points. The gains are larger on non-BN models: for \textbf{ResNet50-GN}, adding STAG to EATA improves accuracy by \textbf{+6.97\%} to 52.3\%, and for \textbf{VitBase-LN}, TENT+STAG achieves the largest boost of \textbf{+11.21\%} points (48.0\% $\rightarrow$ 59.2\%). These results suggest that STAG regularizes adaptation across diverse architectures. We further analyze stability using accuracy running curves under Gaussian noise. As shown in Figure~\ref{fig:running_curves}, STAG (dashed) yields higher accuracy and more stable convergence than the original baselines (solid), with faster initial improvement and a better steady state. Detailed per-condition curves are provided in Figure~\ref{fig:appendix_all_results}.

\paragraph{t-SNE Visualization}
To qualitatively investigate the impact of STAG on the model’s internal representations, we visualize the feature space using t-SNE~\cite{van2008visualizing}. Figure~\ref{fig:tsne} illustrates the feature distributions for VitBase-LN on the ImageNet-C Frost dataset at severity level 5.
As shown in Figure~\ref{fig:tsne}(a), the feature space of the baseline \textbf{TENT} exhibits heavily overlapped clusters, indicating a loss of discriminative power under severe distortions. In contrast, \textbf{TENT+STAG} (Figure~\ref{fig:tsne}(b)) yields significantly tighter and better-separated clusters. This visualization confirms that STAG effectively regularizes the adaptation process, maintaining the semantic structure of the feature space even under significant distribution shifts.

\subsection{Results in Dynamic Scenario}
We evaluate STAG under three realistic dynamic settings proposed by~\cite{sar}, where the test stream is \emph{non-stationary}.

\paragraph{Online Imbalanced Label Distribution Shifts}
This setting simulates \emph{class-prior shift} at test time: the incoming stream is organized into consecutive phases in which the label distribution becomes highly skewed toward a subset of classes, and the dominant classes remain overrepresented for a while before the stream shifts again. Such imbalance can amplify confirmation bias in pseudo-labeling and destabilize entropy-based updates.
As summarized in Table~\ref{tab:label_shifts_bs1}, STAG improves robustness under these imbalanced priors by stabilizing the adaptation signal, yielding average gains of up to \textbf{7.33\%} for TENT (ResNet50-GN) and \textbf{5.79\%} for SAR (VitBase-LN).

\paragraph{Batch Size 1 (Single-sample Adaptation)}
Single-sample adaptation ($B{=}1$) is challenging online regime, where the model updates after each test example with no batch statistics and minimal averaging. This setting is particularly prone to noisy gradients and error accumulation, often causing abrupt performance collapse for unstable objectives.
As shown in Table~\ref{tab:label_shifts_bs1}, STAG acts as a strong regularizer in this extreme regime, providing substantial absolute improvements of \textbf{10.68\%} points for EATA on ResNet50-GN and \textbf{11.71\%} points on VitBase-LN.

\paragraph{Mixed Distribution Shifts}
This setting evaluates adaptation on a mixed-corruption test set, constructed by combining samples from all 15 corruption types. Rather than adapting to a single fixed corruption, the model must handle heterogeneous corruptions within the same evaluation stream, which makes the pseudo-labels and entropy signal more variable.
Table~\ref{tab:mix_shifts_avg} reports the results over a mixture of 15 corruption types. STAG consistently improves performance in this non-stationary stream, achieving a \textbf{1.5\%} points gain for EATA on ResNet50-GN and a \textbf{2.7\%} points gain for SAR on VitBase-LN.

\definecolor{LightPink}{RGB}{255,238,238}


\begin{table}[!t]
\centering
\small
\setlength{\tabcolsep}{3pt}
\renewcommand{\arraystretch}{1.15}
\caption{\textbf{Classification accuracy (\%) on ImageNet-C (severity 5) under the mixed distribution shift.}}
\label{tab:mix_shifts_avg}
\resizebox{\columnwidth}{!}{%
\begin{tabular}{
l|
c >{\columncolor{LightPink}}c c|
c >{\columncolor{LightPink}}c c
}
\hline
Model
& \multicolumn{3}{c|}{ResNet50-GN}
& \multicolumn{3}{c}{VitBase-LN} \\
\hline

Method & Original & + STAG & Gain & Original & + STAG & Gain \\ \hline
        Source & 30.6 & - & - & 29.9 & - & - \\ 
        TENT & 13.3 & - & - & 15.8 & - & - \\ 
        EATA & 38.0 & $\textbf{39.4}_{\pm0.3}$ & +1.5 & 56.5 & $\textbf{57.8}_{\pm0.6}$ & +1.3 \\ 
        SAR & 38.3 & $\textbf{38.9}_{\pm0.5}$ & +0.6 & 57.1 & $\textbf{59.8}_{\pm0.1}$ & +2.7 \\ 
        DeYO & 38.0 & $\textbf{38.8}_{\pm1.6}$ & +0.8 & 58.9 & $\textbf{59.7}_{\pm0.4}$ & +0.8 \\  \hline
    \end{tabular}}
\end{table}
\definecolor{LightPink}{RGB}{255,238,238}

\begin{table*}[t]
\centering
\small
\setlength{\tabcolsep}{3pt}
\renewcommand{\arraystretch}{1.15}
\caption{\textbf{Classification accuracy (\%) on ImageNet-C (severity 5) under the imbalanced label shifts (imbalance ratio = $\infty$) or the batch size 1.}}
\label{tab:label_shifts_bs1}
\resizebox{\textwidth}{!}{%
\begin{tabular}{l|ccc|cccc|cccc|cccc|c}
\multicolumn{1}{l}{} & \multicolumn{3}{c}{Noise} & \multicolumn{4}{c}{Blur} & \multicolumn{4}{c}{Weather} & \multicolumn{4}{c}{Digital} & \\
\textbf{Label Shifts} & Gauss. & Shot & Impul. & Defoc. & Glass & Motion & Zoom & Snow & Frost & Fog & Brit. & Contr. & Elastic & Pixel & JPEG & Avg. \\
\midrule
ResNet50-GN & 18.0 & 19.8 & 17.9 & 19.8 & 11.3 & 21.4 & 24.9 & 40.4 & 47.3 & 33.6 & 69.3 & 36.3 & 18.6 & 28.4 & 52.3 & 30.6 \\
\textbullet\ TENT & 2.6 & 3.2 & 2.9 & 13.5 & 5.4 & 18.8 & 16.7 & 15.8 & 21.1 & 1.8 & 70.4 & 42.1 & 6.3 & 49.4 & 53.6 & 21.6 \\
\textbullet\ EATA & 26.9 & \textbf{31.0} & 26.0 & 18.2 & 17.8 & 23.7 & 26.8 & 34.4 & 32.4 & 41.8 & 66.9 & 33.7 & 26.0 & 41.7 & 41.6 & 32.6 \\
\rowcolor{LightPink} +STAG &
$\textbf{29.6}_{\pm0.3}$ &
$30.8_{\pm1.9}$ &
$\textbf{30.1}_{\pm0.7}$ &
$\textbf{20.2}_{\pm4.6}$ &
$\textbf{21.0}_{\pm1.8}$ &
$\textbf{33.1}_{\pm1.6}$ &
$\textbf{35.4}_{\pm0.9}$ &
$\textbf{46.7}_{\pm0.7}$ &
$\textbf{43.9}_{\pm1.6}$ &
$\textbf{50.4}_{\pm2.5}$ &
$\textbf{68.9}_{\pm0.7}$ &
$\textbf{42.3}_{\pm1.0}$ &
$\textbf{39.6}_{\pm1.6}$ &
$\textbf{53.8}_{\pm0.4}$ &
$\textbf{53.5}_{\pm1.4}$ &
$\textbf{39.9}_{\pm0.4}$ {\scriptsize(+7.33)} \\
\textbullet\ SAR & 33.8 & 36.5 & 34.9 & \textbf{19.3} & \textbf{20.2} & 33.4 & 29.2 & 23.7 & 45.0 & \textbf{34.6} & 71.9 & 46.7 & \textbf{6.9} & 52.1 & 56.3 & 36.3 \\
\rowcolor{LightPink} +STAG &
$\textbf{37.8}_{\pm2.2}$ &
$\textbf{40.4}_{\pm1.8}$ &
$\textbf{39.3}_{\pm1.0}$ &
$18.7_{\pm0.8}$ &
$16.1_{\pm9.7}$ &
$\textbf{35.2}_{\pm2.0}$ &
$\textbf{37.8}_{\pm0.5}$ &
$\textbf{40.4}_{\pm9.0}$ &
$\textbf{47.9}_{\pm0.9}$ &
$18.8_{\pm30.9}$ &
$\textbf{73.2}_{\pm0.1}$ &
$\textbf{49.0}_{\pm0.2}$ &
$1.0_{\pm0.2}$ &
$\textbf{55.1}_{\pm0.1}$ &
$\textbf{56.9}_{\pm0.1}$ &
$\textbf{37.8}_{\pm2.7}$ {\scriptsize(+1.53)} \\
\textbullet\ DeYO & 40.0 & 44.1 & 42.6 & \textbf{22.0} & \textbf{23.5} & 41.1 & 8.2 & 52.2 & 51.2 & 19.5 & 73.2 & 52.7 & \textbf{37.7} & 59.4 & 59.1 & \textbf{41.8} \\
\rowcolor{LightPink} +STAG &
$\textbf{42.6}_{\pm1.2}$ &
$\textbf{44.8}_{\pm0.3}$ &
$\textbf{43.5}_{\pm0.8}$ &
$21.4_{\pm1.2}$ &
$7.7_{\pm11.9}$ &
$\textbf{41.7}_{\pm0.6}$ &
$\textbf{14.3}_{\pm16.0}$ &
$\textbf{54.4}_{\pm1.0}$ &
$\textbf{52.5}_{\pm0.3}$ &
$\textbf{20.4}_{\pm34.1}$ &
$\textbf{73.5}_{\pm0.1}$ &
$\textbf{52.9}_{\pm0.3}$ &
$34.1_{\pm24.1}$ &
$\textbf{60.1}_{\pm0.4}$ &
$\textbf{59.3}_{\pm0.2}$ &
$41.5_{\pm1.8}$ {\scriptsize(-0.22)} \\
\midrule
VitBase-LN & 9.5 & 6.7 & 8.2 & 29.0 & 23.4 & 33.9 & 27.1 & 15.9 & 26.5 & 47.2 & 54.7 & 44.1 & 30.5 & 44.5 & 47.8 & 29.9 \\
\textbullet\ TENT & 32.6 & \textbf{0.9} & 29.3 & 54.6 & 52.5 & 58.2 & 52.8 & 9.4 & 10.4 & 69.3 & 76.1 & 65.9 & 57.7 & 69.4 & 66.4 & 47.0 \\
\rowcolor{LightPink} +STAG &
$\textbf{35.8}_{\pm7.6}$ &
$0.6_{\pm0.6}$ &
$\textbf{38.8}_{\pm16.5}$ &
$\textbf{56.3}_{\pm0.0}$ &
$\textbf{55.7}_{\pm0.1}$ &
$\textbf{60.2}_{\pm0.1}$ &
$\textbf{55.8}_{\pm1.3}$ &
$\textbf{19.1}_{\pm5.0}$ &
$\textbf{15.4}_{\pm3.9}$ &
$\textbf{71.5}_{\pm0.3}$ &
$\textbf{77.1}_{\pm0.1}$ &
$\textbf{67.1}_{\pm0.0}$ &
$\textbf{62.5}_{\pm1.3}$ &
$\textbf{71.2}_{\pm0.1}$ &
$\textbf{68.4}_{\pm0.1}$ &
$\textbf{50.4}_{\pm1.9}$ {\scriptsize(+3.33)} \\
\textbullet\ EATA &
36.3 &
\textbf{37.3} &
\textbf{37.0} &
\textbf{41.3} &
\textbf{44.6} &
\textbf{49.4} &
39.7 &
54.9 &
51.8 &
47.3 &
73.9 &
\textbf{27.5} &
57.6 &
\textbf{64.9} &
62.8 &
48.4 \\
\rowcolor{LightPink} +STAG &
$\textbf{36.9}_{\pm2.1}$ &
$34.7_{\pm1.9}$ &
$36.6_{\pm1.8}$ &
$38.9_{\pm4.7}$ &
$43.8_{\pm1.1}$ &
$49.0_{\pm1.1}$ &
$\textbf{46.1}_{\pm1.8}$ &
$\textbf{55.5}_{\pm1.4}$ &
$\textbf{53.2}_{\pm1.5}$ &
$\textbf{57.9}_{\pm5.0}$ &
$\textbf{74.0}_{\pm0.8}$ &
$21.9_{\pm16.2}$ &
$\textbf{58.9}_{\pm0.8}$ &
$64.8_{\pm1.3}$ &
$\textbf{63.6}_{\pm0.5}$ &
$\textbf{49.1}_{\pm0.6}$ {\scriptsize(+0.64)} \\
\textbullet\ SAR & 46.5 & 36.1 & 48.8 & 55.3 & 54.3 & 58.9 & 54.8 & 51.0 & 46.3 & 69.7 & 76.2 & 66.2 & 60.8 & 69.6 & 66.6 & 57.4 \\
\rowcolor{LightPink} +STAG &
$\textbf{52.7}_{\pm1.6}$ &
$\textbf{51.4}_{\pm3.1}$ &
$\textbf{52.9}_{\pm1.9}$ &
$\textbf{57.8}_{\pm0.1}$ &
$\textbf{59.1}_{\pm0.1}$ &
$\textbf{63.3}_{\pm0.1}$ &
$\textbf{61.4}_{\pm0.6}$ &
$\textbf{67.6}_{\pm0.2}$ &
$\textbf{50.9}_{\pm13.4}$ &
$\textbf{73.7}_{\pm0.1}$ &
$\textbf{77.9}_{\pm0.1}$ &
$\textbf{67.4}_{\pm0.2}$ &
$\textbf{68.5}_{\pm0.3}$ &
$\textbf{73.4}_{\pm0.1}$ &
$\textbf{70.0}_{\pm0.1}$ &
$\textbf{63.2}_{\pm1.3}$ {\scriptsize(+5.79)} \\
\textbullet\ DeYO & 35.8 & 51.7 & 53.4 & 57.7 & 58.8 & 63.7 & 39.0 & 67.9 & 66.2 & 73.1 & 77.9 & 66.6 & 68.9 & 73.6 & 70.3 & 61.6 \\
\rowcolor{LightPink} +STAG &
$\textbf{53.1}_{\pm1.2}$ &
$\textbf{53.6}_{\pm2.6}$ &
$\textbf{54.4}_{\pm1.6}$ &
$\textbf{57.8}_{\pm0.1}$ &
$\textbf{59.0}_{\pm0.2}$ &
$\textbf{64.4}_{\pm0.2}$ &
$\textbf{55.2}_{\pm12.7}$ &
$\textbf{68.6}_{\pm0.1}$ &
$\textbf{66.3}_{\pm0.1}$ &
$\textbf{73.4}_{\pm0.2}$ &
$\textbf{78.0}_{\pm0.0}$ &
$\textbf{66.7}_{\pm0.2}$ &
$\textbf{69.6}_{\pm0.1}$ &
$\textbf{73.9}_{\pm0.1}$ &
$\textbf{70.6}_{\pm0.2}$ &
$\textbf{64.3}_{\pm1.2}$ {\scriptsize(+2.67)} \\
\end{tabular}}

\centering
\small
\setlength{\tabcolsep}{3pt}
\renewcommand{\arraystretch}{1.15}
\resizebox{\textwidth}{!}{%
\begin{tabular}{l|ccc|cccc|cccc|cccc|c}
\multicolumn{1}{l}{} & \multicolumn{3}{c}{Noise} & \multicolumn{4}{c}{Blur} & \multicolumn{4}{c}{Weather} & \multicolumn{4}{c}{Digital} & \\
\textbf{Batch Size 1} & Gauss. & Shot & Impul. & Defoc. & Glass & Motion & Zoom & Snow & Frost & Fog & Brit. & Contr. & Elastic & Pixel & JPEG & Avg. \\
\midrule
ResNet50-GN & 18.0 & 19.8 & 17.9 & 19.8 & 11.3 & 21.4 & 24.9 & 40.4 & 47.3 & 33.6 & 69.3 & 36.3 & 18.6 & 28.4 & 52.3 & 30.6 \\

\textbullet\ TENT
& 2.3 & 3.1 & 2.7 & 13.2 & 4.7 & 18.0 & 18.7 & 16.2 & 20.8 & 1.4 & 70.3 & 42.2 & 6.0 & 49.3 & 53.7 & 21.5 \\

\textbullet\ EATA
& 24.9 & 28.1 & 25.6 & 17.9 & 17.1 & 28.6 & 29.4 & 44.7 & 44.4 & 39.8 & 70.9 & 44.3 & 26.9 & 46.4 & 55.6 & 36.3 \\

\rowcolor{LightPink} +STAG
& $\textbf{38.9}_{\pm0.3}$ & $\textbf{42.2}_{\pm0.3}$ & $\textbf{40.7}_{\pm0.2}$
& $\textbf{17.9}_{\pm0.1}$ & $\textbf{27.3}_{\pm0.4}$ & $\textbf{39.9}_{\pm0.4}$ & $\textbf{43.0}_{\pm0.3}$
& $\textbf{55.1}_{\pm0.3}$ & $\textbf{51.9}_{\pm0.2}$ & $\textbf{58.4}_{\pm0.1}$ & $\textbf{73.2}_{\pm0.1}$
& $\textbf{51.8}_{\pm0.2}$ & $\textbf{47.0}_{\pm0.2}$ & $\textbf{58.6}_{\pm0.2}$ & $\textbf{59.0}_{\pm0.2}$
& $\textbf{47.0}_{\pm0.1}$ {\scriptsize(+10.68)} \\

\textbullet\ SAR
& 23.4 & 26.6 & 23.6 & \textbf{18.6} & \textbf{15.0} & 28.8 & 30.3 & 45.1 & 44.6 & 27.6 & 72.3 & 44.7 & \textbf{13.2} & 46.8 & 56.1 & 34.4 \\

\rowcolor{LightPink} +STAG
& $\textbf{43.5}_{\pm0.3}$ & $\textbf{46.0}_{\pm0.2}$ & $\textbf{44.7}_{\pm0.1}$
& $12.6_{\pm0.1}$ & $10.2_{\pm1.9}$ & $\textbf{41.6}_{\pm0.3}$ & $\textbf{46.5}_{\pm0.4}$
& $\textbf{59.0}_{\pm0.3}$ & $\textbf{54.0}_{\pm0.0}$ & $\textbf{51.9}_{\pm19.0}$ & $\textbf{73.2}_{\pm0.1}$
& $\textbf{52.6}_{\pm0.1}$ & $6.5_{\pm0.8}$ & $\textbf{61.8}_{\pm0.1}$ & $\textbf{58.4}_{\pm0.0}$
& $\textbf{44.2}_{\pm1.1}$ {\scriptsize(+9.72)} \\

\textbullet\ DeYO
& 37.1 & 40.7 & 39.0 & \textbf{21.3} & 23.1 & 37.4 & 37.3 & 51.6 & 49.5 & 3.4 & 73.2 & 50.7 & \textbf{41.0} & 56.4 & 58.2 & 41.3 \\

\rowcolor{LightPink} +STAG
& $\textbf{41.2}_{\pm0.3}$ & $\textbf{44.3}_{\pm0.1}$ & $\textbf{42.5}_{\pm0.4}$
& $20.8_{\pm0.2}$ & $\textbf{24.3}_{\pm0.3}$ & $\textbf{40.1}_{\pm0.1}$ & $\textbf{40.5}_{\pm0.7}$
& $\textbf{56.3}_{\pm0.2}$ & $\textbf{52.3}_{\pm0.1}$ & $\textbf{58.7}_{\pm0.5}$ & $\textbf{73.7}_{\pm0.0}$
& $\textbf{52.4}_{\pm0.1}$ & $27.1_{\pm15.5}$ & $\textbf{59.5}_{\pm0.1}$ & $\textbf{59.0}_{\pm0.1}$
& $\textbf{46.2}_{\pm1.0}$ {\scriptsize(+4.86)} \\

\midrule
VitBase-LN & 9.5 & 6.7 & 8.2 & 29.0 & 23.4 & 33.9 & 27.1 & 15.9 & 26.5 & 47.2 & 54.7 & 44.1 & 30.5 & 44.5 & 47.8 & 29.9 \\

\textbullet\ TENT
& 42.3 & 1.0 & \textbf{43.2} & 52.6 & 48.1 & 55.4 & 50.1 & 16.5 & 18.4 & 66.5 & 74.9 & 64.7 & 52.6 & 66.8 & 64.1 & 47.8 \\

\rowcolor{LightPink} +STAG
& $\textbf{53.6}_{\pm0.4}$ & $\textbf{18.4}_{\pm31.3}$ & $38.3_{\pm28.5}$
& $\textbf{58.3}_{\pm0.2}$ & $\textbf{59.1}_{\pm0.0}$ & $\textbf{63.9}_{\pm0.0}$ & $\textbf{62.2}_{\pm0.2}$
& $\textbf{46.0}_{\pm38.0}$ & $\textbf{45.2}_{\pm36.6}$ & $\textbf{73.3}_{\pm0.1}$ & $\textbf{77.5}_{\pm0.0}$
& $\textbf{68.2}_{\pm0.1}$ & $\textbf{68.7}_{\pm0.1}$ & $\textbf{72.9}_{\pm0.1}$ & $\textbf{69.9}_{\pm0.1}$
& $\textbf{58.4}_{\pm5.1}$ {\scriptsize(+10.55)} \\

\textbullet\ EATA
& 32.6 & 27.7 & 33.7 & 44.9 & 40.8 & 47.3 & 42.7 & 39.4 & 43.8 & 62.4 & 67.0 & \textbf{62.0} & 47.3 & 60.8 & 59.3 & 47.4 \\

\rowcolor{LightPink} +STAG
& $\textbf{50.7}_{\pm1.4}$ & $\textbf{50.3}_{\pm1.8}$ & $\textbf{51.4}_{\pm1.6}$
& $\textbf{53.7}_{\pm0.9}$ & $\textbf{55.5}_{\pm1.1}$ & $\textbf{60.0}_{\pm1.0}$ & $\textbf{59.0}_{\pm1.3}$
& $\textbf{64.0}_{\pm1.8}$ & $\textbf{62.5}_{\pm1.6}$ & $\textbf{71.6}_{\pm0.4}$ & $\textbf{75.3}_{\pm0.8}$
& $46.0_{\pm34.3}$ & $\textbf{49.6}_{\pm29.1}$ & $\textbf{70.2}_{\pm0.9}$ & $\textbf{67.5}_{\pm0.7}$
& $\textbf{59.2}_{\pm2.7}$ {\scriptsize(+11.71)} \\

\textbullet\ SAR
& 40.8 & 36.5 & 41.8 & 53.5 & 50.6 & 57.4 & 52.8 & 58.8 & 54.8 & 67.6 & 75.5 & 65.6 & 58.2 & 68.8 & 65.8 & 56.6 \\

\rowcolor{LightPink} +STAG
& $\textbf{56.3}_{\pm0.2}$ & $\textbf{57.6}_{\pm0.1}$ & $\textbf{57.1}_{\pm0.1}$
& $\textbf{59.2}_{\pm0.2}$ & $\textbf{60.5}_{\pm0.1}$ & $\textbf{65.9}_{\pm0.2}$ & $\textbf{65.1}_{\pm1.9}$
& $\textbf{67.9}_{\pm1.5}$ & $\textbf{67.2}_{\pm0.2}$ & $\textbf{72.5}_{\pm0.5}$ & $\textbf{76.8}_{\pm0.2}$
& $\textbf{67.6}_{\pm0.4}$ & $\textbf{68.0}_{\pm1.8}$ & $\textbf{72.2}_{\pm1.0}$ & $\textbf{69.1}_{\pm0.1}$
& $\textbf{65.5}_{\pm0.3}$ {\scriptsize(+8.95)} \\

\textbullet\ DeYO
& 51.0 & 45.8 & 51.8 & 57.7 & 57.6 & \textbf{62.4} & 57.0 & 66.2 & 65.1 & 72.8 & \textbf{77.9} & \textbf{68.0} & 66.4 & 72.5 & 69.6 & 62.8 \\

\rowcolor{LightPink} +STAG
& $\textbf{56.2}_{\pm0.0}$ & $\textbf{57.6}_{\pm0.1}$ & $\textbf{56.9}_{\pm0.1}$
& $\textbf{58.9}_{\pm0.2}$ & $\textbf{60.3}_{\pm0.1}$ & $44.9_{\pm36.1}$ & $\textbf{61.5}_{\pm7.8}$
& $\textbf{69.2}_{\pm0.0}$ & $\textbf{66.9}_{\pm0.5}$ & $\textbf{73.4}_{\pm0.1}$ & $77.0_{\pm0.1}$
& $49.8_{\pm29.6}$ & $\textbf{70.5}_{\pm0.1}$ & $\textbf{73.4}_{\pm0.1}$ & $\textbf{70.4}_{\pm0.1}$
& $\textbf{63.1}_{\pm2.5}$ {\scriptsize(+0.34)} \\

\end{tabular}}

\end{table*}

\subsection{Results in Continual Test-Time Adaptation}
Continual TTA considers long-horizon online adaptation on a \emph{non-stationary} target stream without access to source data, requiring robustness to error accumulation and catastrophic forgetting. For classification, models pre-trained on clean CIFAR-10 or CIFAR-100 are continually adapted online to sequentially encountered corruption types. All results are reported at the highest corruption severity (level 5). For segmentation, following~\cite{wang2022cotta}, we evaluate Cityscapes-to-ACDC under a cyclic weather stream that repeats Fog $\rightarrow$ Night $\rightarrow$ Rain $\rightarrow$ Snow for ten rounds (40 stages), emphasizing long-term stability through repeated re-adaptation to previously encountered conditions.
\vspace{-2mm}
\paragraph{Continual classification.}
Table~\ref{tab:continual_tta} reports classification error rates (\%) on CIFAR-10-C and CIFAR-100-C (severity 5). STAG consistently reduces error across diverse continual TTA baselines, yielding substantial gains for BN Adapt (\textbf{20.4$\rightarrow$17.9}, \textbf{39.3$\rightarrow$36.7}) and TENT (\textbf{19.0$\rightarrow$17.4}, \textbf{37.2$\rightarrow$35.4}), while also providing additional improvements for stronger methods such as CoTTA and EcoTTA. The error rates for each corruption type are provided in Appendix~\ref{sec:append_add} (Tables~\ref{tab:cifar-10-c_continual} and~\ref{tab:cifar-100-c_continual}). These results indicate that STAG complements existing continual objectives as a lightweight stabilizer, improving robustness without changing each method's optimization protocol.

\begin{figure}[t]
    \centering
    \includegraphics[width=\columnwidth]{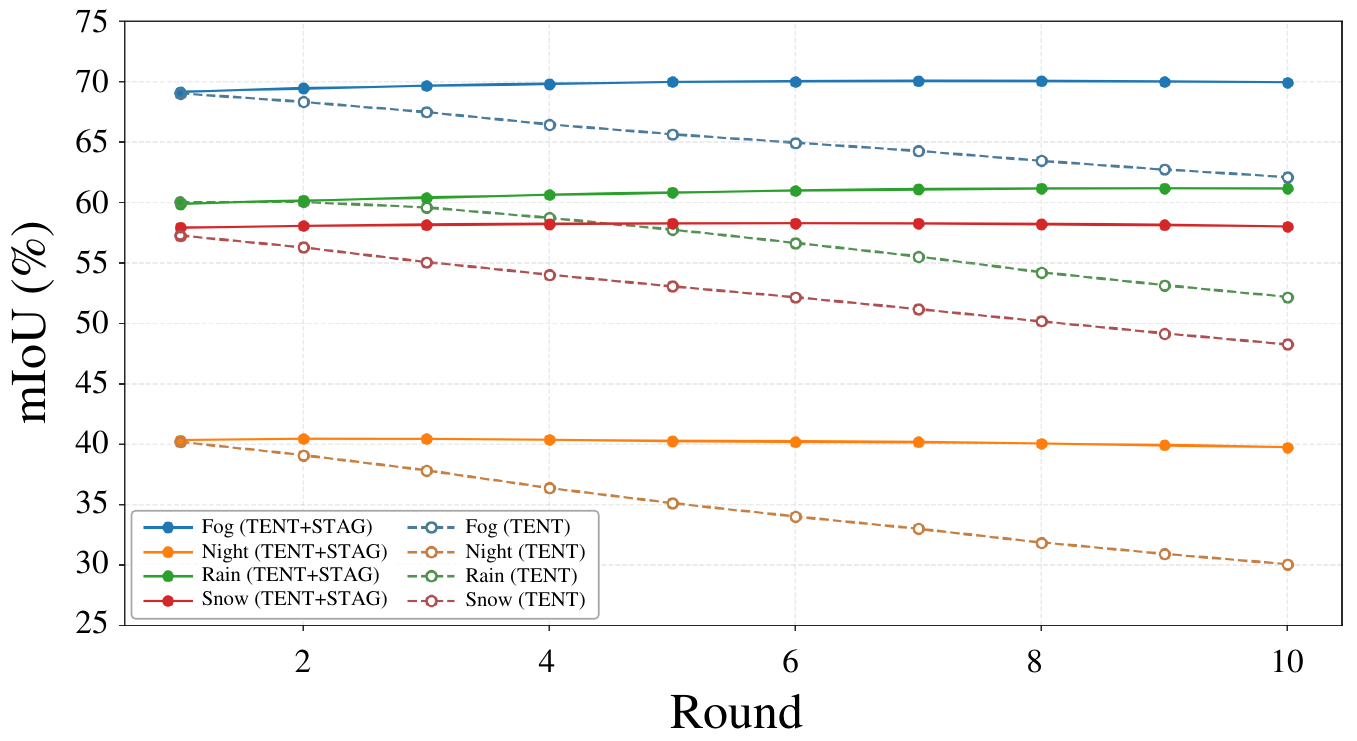}
    \vspace{-3mm}
    \caption{\textbf{Semantic segmentation performance (mIoU, \%)} between TENT and its STAG-integrated version on the Cityscapes-to-ACDC continual test-time adaptation tasks.}
    \label{fig:continual_graph}
    \vspace{-2mm}
\end{figure}

\begin{figure}[t]
    \centering
    \includegraphics[width=\columnwidth]{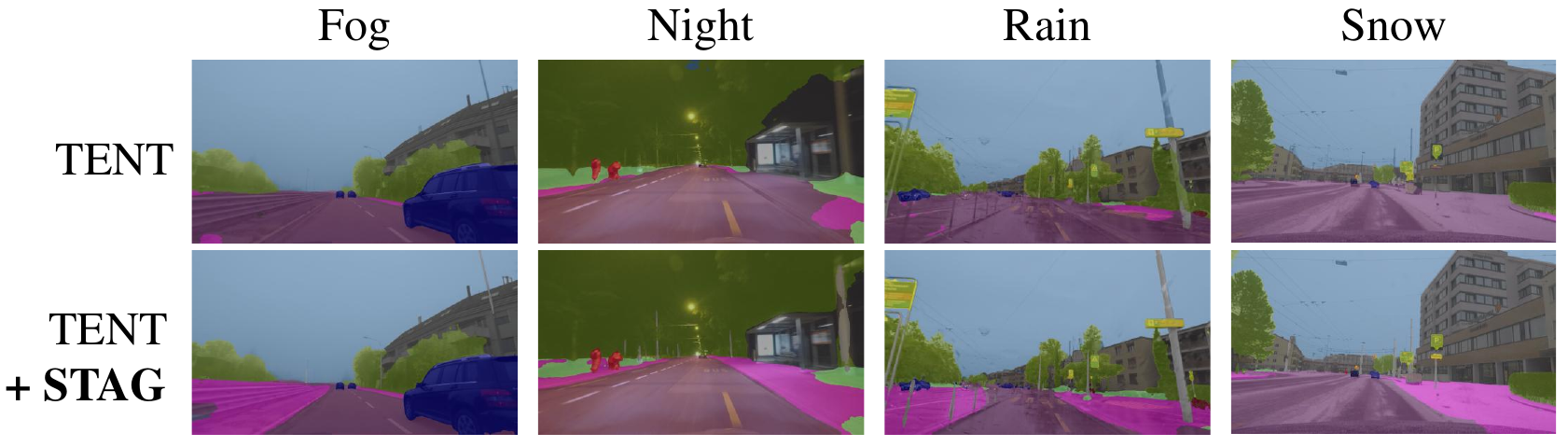}
    \vspace{-3mm}
    \caption{\textbf{Semantic segmentation maps} on the Cityscapes-to-ACDC continual test-time adaptation task at the 10th round.}
    \label{fig:continual_segmap}
    \vspace{-3mm}
\end{figure}

\paragraph{Continual semantic segmentation.}
Figure~\ref{fig:continual_graph} compares TENT and TENT+STAG on Cityscapes-to-ACDC over ten rounds using per-weather mIoU. TENT exhibits clear forgetting, with performance progressively degrading across cycles due to accumulated drift from sequential adaptation. In contrast, TENT+STAG maintains consistently higher mIoU across repeated cycles and, in some cases, even shows performance improvements, indicating that STAG provides a structural prior that counteracts long-horizon drift in continual adaptation. Detailed results, including BN Adapt and BN Adapt+STAG, are reported in Appendix ~\ref{sec:append_add} (Table~\ref{tab:acdc_continual}). Figure~\ref{fig:continual_segmap} qualitatively supports this observation, showing that TENT+STAG produces more coherent segmentation maps with fewer artifacts and better scene preservation under Fog, Night, Rain, and Snow. These results demonstrate that STAG mitigates forgetting and stabilizes continual TTA under cyclic, non-stationary shifts.

\begin{table}
\centering
\small
\setlength{\tabcolsep}{3pt}
\renewcommand{\arraystretch}{1.15}
\caption{\textbf{Classification error rate (\%) on CIFAR-10-C (severity 5) and CIFAR-100-C (severity 5) in continual test-time adaptation tasks.}}
\label{tab:continual_tta}
\resizebox{\columnwidth}{!}{%
\begin{tabular}{
l|
c >{\columncolor{LightPink}}c c|
c >{\columncolor{LightPink}}c c
}
\hline
Dataset (Model)
& \multicolumn{3}{c|}{CIFAR-10-C (WRN-28)}
& \multicolumn{3}{c}{CIFAR-100-C (WRN-40)} \\
\hline

Method & Original & + STAG & Gain & Original & + STAG & Gain \\ \hline
        Source & 43.5 & - & - & 46.8 & - & - \\ 
        BN Adapt & 20.4 & $\textbf{17.9}_{\pm0.0}$ & +2.5 & 39.3 & $\textbf{36.7}_{\pm0.0}$ & +2.6 \\ 
        TENT & 19.0 & $\textbf{17.4}_{\pm0.0}$ & +1.6 & 37.2 & $\textbf{35.4}_{\pm0.0}$ & +1.8 \\ 
        CoTTA & 16.3 & $\textbf{16.1}_{\pm0.0}$ & +0.2 & 38.1 & $\textbf{37.4}_{\pm0.0}$ & +0.7 \\ 
        EcoTTA {\scriptsize ($\mathrm{K}=4$)} & 17.9 & $\textbf{17.5}_{\pm0.0}$ & +0.4 & 36.8 & $\textbf{36.6}_{\pm0.0}$ & +0.2 \\ 
        EcoTTA {\scriptsize ($\mathrm{K}=5$)} & 18.0 & $\textbf{17.6}_{\pm0.0}$ & +0.4 & 36.8 & $\textbf{36.7}_{\pm0.0}$ & +0.1 \\ \hline
    \end{tabular}}
\vspace{-5mm}
\end{table}

\subsection{Ablation Study}
\begin{figure}[t]
    \centering
    \includegraphics[width=\columnwidth]{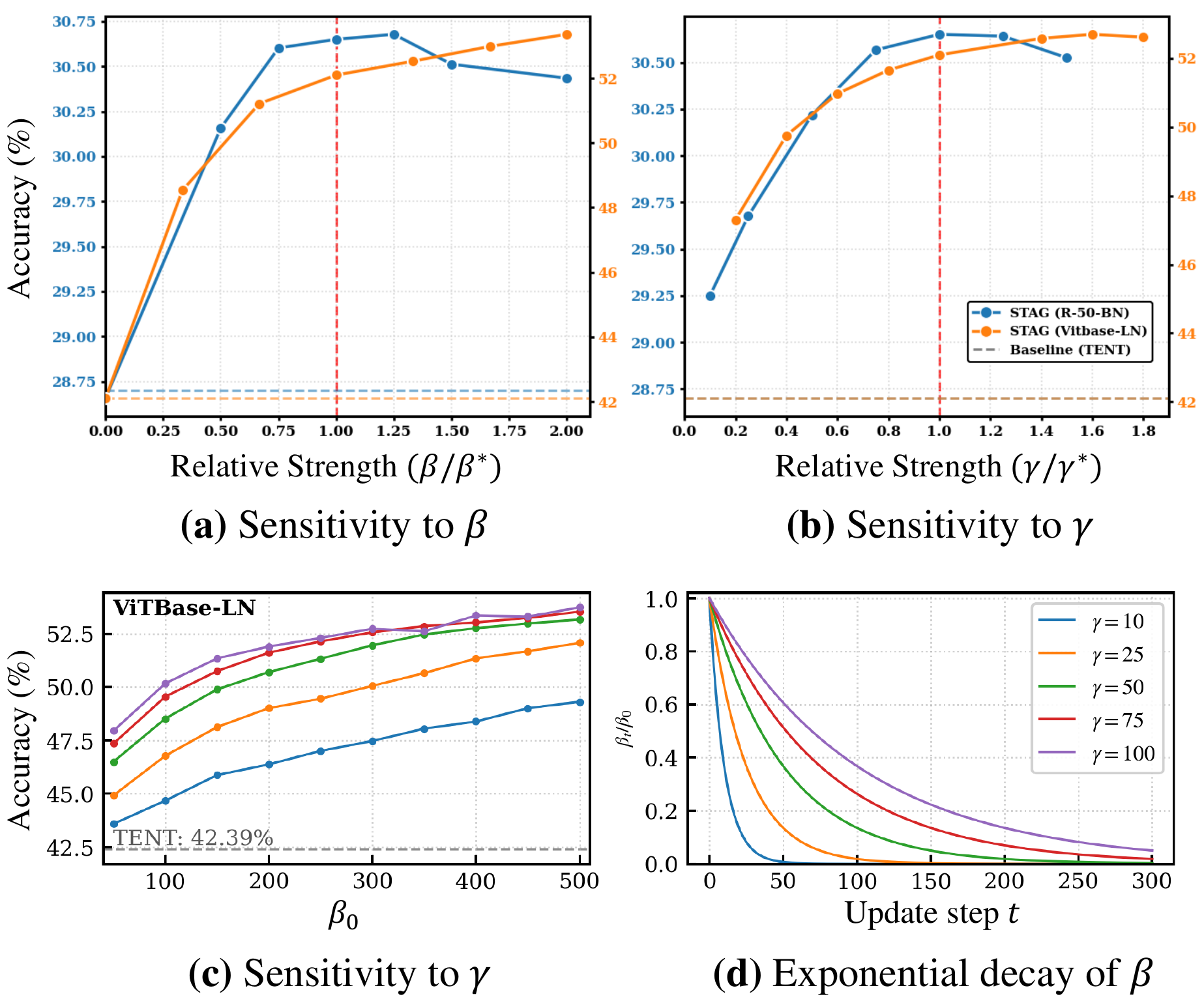}
    \vspace{-2mm}
    \caption{\textbf{Hyper-parameter sensitivity analysis} of STAG on ImageNet-C Gaussian Noise (Severity 5).}
    \label{fig:hp_sensitivity}
    \vspace{-0mm}
\end{figure}

\paragraph{Sensitivity of Hyper-parameters}
We analyze the sensitivity of STAG to its two key hyper-parameters: the initial balancing coefficient $\beta_0$ and the decay factor $\gamma$. Figure~\ref{fig:hp_sensitivity} reports Top-1 accuracy on ImageNet-C Gaussian noise (severity 5), where the x-axes in (a) and (b) are normalized by the optimal values ($\beta^*,\gamma^*$; see Appendix ~\ref{sec:append_add} (Table ~\ref{tab:appendix_hp})) . The red dashed line indicates the normalized optimum (1.0), around which we observe a broad plateau, suggesting that STAG is robust to moderate perturbations of both hyper-parameters. Across a wide range, STAG remains above the TENT baseline and shows similar trends on ResNet50-BN and VitBase-LN. Panel (c) shows that larger $\beta_0$ and slower decay (larger $\gamma$) generally improve performance on VitBase-LN, while (d) illustrates the corresponding schedule $\beta_t/\beta_0=\exp(-t/\gamma)$. Overall, STAG provides robust gains without requiring precise hyper-parameter tuning.

\begin{table}[t]
\centering
\caption{\textbf{Ablation study on method design} using VitBase-LN on ImageNet-C Gaussian Noise (Severity 5).}
\label{tab:ablation_design}
\resizebox{\columnwidth}{!}{%
\begin{tabular}{lcccc}
\toprule
\textbf{Design Variant} & \textbf{TENT} & \textbf{EATA} & \textbf{SAR} & \textbf{DeYO} \\ \midrule
Baseline (No STAG) & 42.17 & 51.07 & 44.06 & 53.11 \\ \midrule
Full / Soft & 29.80 & 55.08 & 52.90 & 50.32 \\
Full / Hard & 49.24 & 52.17 & 47.92 & 51.86 \\
Diagonal / Soft & \textbf{52.92} & 54.79 & 52.24 & 48.43 \\
\rowcolor{LightPink} \textbf{Diagonal / Hard (Ours)} & 52.13 & \textbf{55.26} & \textbf{53.87} & \textbf{55.88} \\ 
\bottomrule
\end{tabular}%
}
\end{table}

\vspace{-2mm}
\paragraph{Ablation on Design Choices}
We conduct an ablation study on VitBase-LN under Gaussian noise to validate our design choices for structural anchor (\textit{Diagonal vs. Full}) and weighting schemes (\textit{Hard vs. Soft labels}). Table~\ref{tab:ablation_design} indicates that the \textbf{Diagonal/Hard} setting performs best in most cases across the baselines. Using only diagonal gradient elements outperforms the full gradient (e.g., 52.13\% vs. 49.24\% for TENT), indicating that the diagonal approximation captures useful structure while mitigating noise from cross-parameter correlations. Hard pseudo-labels also outperform soft labels, which can become ambiguous under heavy noise. With this design, STAG improves TENT and SAR by \textbf{+9.96\%} points and \textbf{+9.81\%} points, respectively.
\vspace{-2mm}
\paragraph{Computational Efficiency}
Empirically, comparing forward/backward cost, runtime, and memory usage shows that STAG introduces little to no additional overhead across baselines. This is because STAG provides structural guidance in closed form, without requiring extra backpropagation beyond the underlying baseline. Detailed statistics are reported in Appendix~\ref{sec:append_add} (Table~\ref{tab:efficiency}).

\section{Conclusion}
\label{sec:conclusion}
We presented \textbf{STAG} (\textbf{S}tructural \textbf{T}est-time \textbf{A}lignment of \textbf{G}radients), a lightweight, \textbf{plug-and-play} regularizer for online test-time adaptation under extreme information scarcity. STAG exploits an always-available structural signal—the classifier’s intrinsic geometry—by deriving class-wise structural anchors via self-structural entropy and aligning predicted-class entropy gradients to these anchors during adaptation. Thanks to its analytical closed-form formulation, STAG introduces \textbf{near-zero} memory and latency overhead and requires no additional backpropagation beyond the underlying baseline, making it practical for real-time deployment. Extensive experiments across corrupted image classification and continual semantic segmentation show that STAG consistently strengthens a wide range of TTA baselines across both CNN and Transformer architectures, while improving robustness in challenging \emph{in-the-wild} settings such as single-sample adaptation, imbalanced label shifts, and long-horizon continual adaptation. Future work includes extending structural alignment to richer heads and broader prediction settings, and developing deeper theoretical characterizations of when and why classifier geometry provides reliable guidance for test-time learning.

\section*{Impact Statement}
This paper proposes a lightweight, plug-and-play regularizer for test-time adaptation that improves robustness to distribution shift without additional supervision or extra backpropagation. Its main positive impact is more reliable deployment of vision models under real-world corruptions (e.g., adverse weather), reducing performance degradation in practice. Potential risks are consistent with adaptive systems in general: online updates can still fail under unexpected out-of-distribution inputs or accumulate errors over long horizons, which may be problematic in safety-critical settings. We mitigate these concerns by evaluating across diverse shifts, architectures, and challenging continual regimes, and we recommend monitoring and conservative deployment when used in high-stakes applications.

\bibliography{ICML_2026}

@inproceedings{wang2020tent,
  title={Tent: Fully Test-Time Adaptation by Entropy Minimization},
  author={Wang, Dequan and Shelhamer, Evan and Liu, Shaoteng and Olshausen, Bruno and Darrell, Trevor},
  booktitle={International Conference on Learning Representations},
  year={2021}
}

@inproceedings{zagoruyko2016wide,
  title={Wide Residual Networks},
  author={Zagoruyko, Sergey and Komodakis, Nikos},
  booktitle={British Machine Vision Conference},
  year={2016}
}

@article{schneider2020improving,
  title={Improving robustness against common corruptions by covariate shift adaptation},
  author={Schneider, Steffen and Rusak, Evgenia and Eck, Luisa and Bringmann, Oliver and Brendel, Wieland and Bethge, Matthias},
  journal={Advances in Neural Information Processing Systems},
  volume={33},
  pages={11539--11551},
  year={2020}
}

@inproceedings{eata,
  title={Efficient test-time model adaptation without forgetting},
  author={Niu, Shuaicheng and Wu, Jiaxiang and Zhang, Yifan and Chen, Yaofo and Zheng, Shijian and Zhao, Peilin and Tan, Mingkui},
  booktitle={International conference on machine learning},
  pages={16888--16905},
  year={2022},
  organization={PMLR}
}

@inproceedings{
sar,
title={Towards Stable Test-time Adaptation in Dynamic Wild World},
author={Shuaicheng Niu and Jiaxiang Wu and Yifan Zhang and Zhiquan Wen and Yaofo Chen and Peilin Zhao and Mingkui Tan},
booktitle={International Conference on Learning Representations},
year={2023}
}

@inproceedings{
hendrycks2019benchmarking,
title={Benchmarking Neural Network Robustness to Common Corruptions and Perturbations},
author={Dan Hendrycks and Thomas Dietterich},
booktitle={International Conference on Learning Representations},
year={2019}
}

@inproceedings{shot,
  title={Do we really need to access the source data? source hypothesis transfer for unsupervised domain adaptation},
  author={Liang, Jian and Hu, Dapeng and Feng, Jiashi},
  booktitle={International Conference on Machine Learning},
  pages={6028--6039},
  year={2020},
  organization={PMLR}
}

@article{krizhevsky2017imagenet,
  title={Imagenet classification with deep convolutional neural networks},
  author={Krizhevsky, Alex and Sutskever, Ilya and Hinton, Geoffrey E},
  journal={Communications of the ACM},
  volume={60},
  number={6},
  pages={84--90},
  year={2017},
  publisher={AcM New York, NY, USA}
}

@inproceedings{lee2022confidence,
  title={Confidence score for source-free unsupervised domain adaptation},
  author={Lee, Jonghyun and Jung, Dahuin and Yim, Junho and Yoon, Sungroh},
  booktitle={International Conference on Machine Learning},
  pages={12365--12377},
  year={2022},
  organization={PMLR}
}

@article{csurka2017domain,
  title={Domain adaptation for visual applications: A comprehensive survey},
  author={Csurka, Gabriela},
  journal={arXiv preprint arXiv:1702.05374},
  year={2017}
}

@inproceedings{muandet2013domain,
  title={Domain generalization via invariant feature representation},
  author={Muandet, Krikamol and Balduzzi, David and Sch{\"o}lkopf, Bernhard},
  booktitle={International conference on machine learning},
  pages={10--18},
  year={2013},
  organization={PMLR}
}

@inproceedings{ganin2015unsupervised,
  title={Unsupervised domain adaptation by backpropagation},
  author={Ganin, Yaroslav and Lempitsky, Victor},
  booktitle={International conference on machine learning},
  pages={1180--1189},
  year={2015},
  organization={PMLR}
}

@article{van2008visualizing,
  title={Visualizing data using t-SNE.},
  author={Van der Maaten, Laurens and Hinton, Geoffrey},
  journal={Journal of machine learning research},
  volume={9},
  number={11},
  year={2008}
}

@inproceedings{deyo,
  title={Entropy is not enough for test-time adaptation: From the perspective of disentangled factors},
  author={Lee, Jonghyun and Jung, Dahuin and Lee, Saehyung and Park, Junsung and Shin, Juhyeon and Hwang, Uiwon and Yoon, Sungroh},
  booktitle={International Conference on Learning Representations},
  year={2024}
}

@article{xie2021segformer,
  title={SegFormer: Simple and efficient design for semantic segmentation with transformers},
  author={Xie, Enze and Wang, Wenhai and Yu, Zhiding and Anandkumar, Anima and Alvarez, Jose M and Luo, Ping},
  journal={Advances in neural information processing systems},
  volume={34},
  pages={12077--12090},
  year={2021}
}

@misc{rw2019timm,
  author = {Ross Wightman},
  title = {PyTorch Image Models},
  year = {2019},
  publisher = {GitHub},
  journal = {GitHub repository},
  doi = {10.5281/zenodo.4414861},
  howpublished = {\url{https://github.com/rwightman/pytorch-image-models}}
}

@inproceedings{wang2022cotta,
  title={Continual test-time domain adaptation},
  author={Wang, Qin and Fink, Olga and Van Gool, Luc and Dai, Dengxin},
  booktitle={Proceedings of the IEEE/CVF Conference on Computer Vision and Pattern Recognition},
  pages={7201--7211},
  year={2022}
}

@inproceedings{song2023ecotta,
  title={Ecotta: Memory-efficient continual test-time adaptation via self-distilled regularization},
  author={Song, Junha and Lee, Jungsoo and Kweon, In So and Choi, Sungha},
  booktitle={Proceedings of the IEEE/CVF Conference on Computer Vision and Pattern Recognition},
  pages={11920--11929},
  year={2023}
}

@inproceedings{cordts2016cityscapes,
  title={The cityscapes dataset for semantic urban scene understanding},
  author={Cordts, Marius and Omran, Mohamed and Ramos, Sebastian and Rehfeld, Timo and Enzweiler, Markus and Benenson, Rodrigo and Franke, Uwe and Roth, Stefan and Schiele, Bernt},
  booktitle={Proceedings of the IEEE conference on computer vision and pattern recognition},
  pages={3213--3223},
  year={2016}
}

@inproceedings{sakaridis2021acdc,
  title={ACDC: The adverse conditions dataset with correspondences for semantic driving scene understanding},
  author={Sakaridis, Christos and Dai, Dengxin and Van Gool, Luc},
  booktitle={Proceedings of the IEEE/CVF international conference on computer vision},
  pages={10765--10775},
  year={2021}
}

@article{bn,
  title={Evaluating prediction-time batch normalization for robustness under covariate shift},
  author={Nado, Zachary and Padhy, Shreyas and Sculley, D and D'Amour, Alexander and Lakshminarayanan, Balaji and Snoek, Jasper},
  journal={arXiv preprint arXiv:2006.10963},
  year={2020}
}

@inproceedings{
hwang2024sf,
title={{SF}({DA})$^2$: Source-free Domain Adaptation Through the Lens of Data Augmentation},
author={Uiwon Hwang and Jonghyun Lee and Juhyeon Shin and Sungroh Yoon},
booktitle={International Conference on Learning Representations},
year={2024}
}
\bibliographystyle{icml2026}

\newpage
\appendix
\onecolumn
\section{Analytical Derivation and Implementation-Consistent Formulation}
\label{sec:appendix_derivation}

This appendix provides an analytical derivation of the Shannon entropy gradient used in our method and clarifies
the exact quantities employed in \ref{sec:method} (Anchor and Sample-wise Gradient Extraction).
All logarithms are natural.

\subsection{Setup and Notation}
Let $\mathbf{h} = f_{\bm{\phi}}(\mathbf{x}) \in \mathbb{R}^D$ be the feature representation of an input $\mathbf{x}$.
The logit $z_i$ for class $i$ is
\begin{equation}
z_i = \mathbf{w}_i^\top \mathbf{h} + b_i,
\end{equation}
and the predictive distribution is $p_i = \sigma(\mathbf{z})_i$ via softmax:
\begin{equation}
p_i = \frac{\exp(z_i)}{\sum_{j=1}^c \exp(z_j)}.
\end{equation}
The Shannon entropy loss is
\begin{equation}
\mathcal{L}_{\text{ent}}(\mathbf{x}) = H(\mathbf{p}) = -\sum_{i=1}^c p_i \ln p_i.
\end{equation}

\subsection{Entropy Gradient with Respect to Logits}
\paragraph{Step 1: Softmax Jacobian.}
The derivative of $p_i$ with respect to $z_j$ is
\begin{equation}
\frac{\partial p_i}{\partial z_j} = p_i(\delta_{ij}-p_j),
\end{equation}
where $\delta_{ij}$ is the Kronecker delta.

\paragraph{Step 2: Entropy derivative with respect to probabilities.}
\begin{equation}
\frac{\partial H}{\partial p_i} = -(\ln p_i + 1).
\end{equation}

\paragraph{Step 3: Chain rule to logits.}
\begin{align}
\frac{\partial H}{\partial z_j}
&= \sum_{i=1}^c \frac{\partial H}{\partial p_i}\frac{\partial p_i}{\partial z_j}
= -\sum_{i=1}^c (\ln p_i + 1)\, p_i(\delta_{ij}-p_j) \nonumber \\
&= -(\ln p_j + 1)p_j + p_j \sum_{i=1}^c (\ln p_i + 1)p_i \nonumber \\
&= -(\ln p_j + 1)p_j + p_j\left(\sum_{i=1}^c p_i\ln p_i + \sum_{i=1}^c p_i\right) \nonumber \\
&= -p_j\ln p_j - p_j + p_j\left(-H(\mathbf{p}) + 1\right) \nonumber \\
&= -p_j\left(\ln p_j + H(\mathbf{p})\right).
\label{eq:appendix_dH_dz}
\end{align}

\subsection{Entropy Gradient with Respect to Classifier Weights (Sample-wise Gradient Extraction)}
\label{sec:appendix_phase2}
Given an input $\mathbf{x}$ with feature $\mathbf{h}=f_{\bm{\phi}}(\mathbf{x})\in\mathbb{R}^D$, we compute the \textit{logit-level sensitivity} $s$ for class $k$ as the derivative of the entropy with respect to the $k$-th logit:
\begin{equation}
s \triangleq \frac{\partial H(\mathbf{p})}{\partial z_k} = -p_k\big(\ln p_k + H(\mathbf{p})\big).
\label{eq:appendix_logit_sensitivity}
\end{equation}
The resulting entropy gradient with respect to the class weight $\mathbf{w}_k$ is then $\nabla_{\mathbf{w}_k}\mathcal{L}_{\text{ent}}(\mathbf{x}) = s \cdot \mathbf{h}$.

For a test sample $\mathbf{x}_t$ with feature $\mathbf{h}_t$ and pseudo-label $\hat{y}_t=\arg\max_k p_{t,k}$, we define the sample-wise direction in the weight geometry as:
\begin{equation}
\mathbf{g}_t \triangleq s_t \cdot \mathbf{h}_t, \quad \text{where} \quad s_t = -p_{t,\hat{y}_t}\big(\ln p_{t,\hat{y}_t} + H(\mathbf{p}_t)\big).
\label{eq:appendix_gt}
\end{equation}
This $s_t$ captures the instantaneous sensitivity of the model's prediction at the logit level for each test instance.

\subsection{Self-Structural Entropy and Diagonal Structural Sensitivity (Anchor Extraction)}
This section defines the self-structural quantities used for anchor extraction.
From the classifier weights $W=[\mathbf{w}_1;\ldots;\mathbf{w}_c]\in\mathbb{R}^{c\times D}$, we construct the
self-structural similarity logits
\begin{equation}
Z \in \mathbb{R}^{c\times c}, \qquad 
Z_{kj} = \mathbf{w}_j^\top \mathbf{w}_k + b_j,
\end{equation}
and apply row-wise softmax to obtain $p_{kj}=\mathrm{softmax}(Z_k)_j$.
The self-structural entropy for each row $k$ is
\begin{equation}
H_k \triangleq -\sum_{j=1}^c p_{kj}\ln p_{kj}.
\end{equation}
By Eq.~\eqref{eq:appendix_dH_dz}, the analytical derivative of $H_k$ with respect to the self-structural logits is
\begin{equation}
\frac{\partial H_k}{\partial Z_{kj}} = -p_{kj}\big(\ln p_{kj} + H_k\big).
\label{eq:appendix_dHk_dZ}
\end{equation}
To define the anchor as the direction that minimizes structural ambiguity, we utilize the diagonal component of the entropy gradient (Eq.~\ref{eq:appendix_dH_dz}) as the scalar \textit{structural sensitivity} $s_k$:
\begin{equation}
s_k \triangleq -p_{kk}\big(\ln p_{kk} + H_k\big).
\label{eq:appendix_sk}
\end{equation}
Using this sensitivity, the structural anchor for class $k$ is formulated as:
\begin{equation}
\mathbf{a}_k = s_k \mathbf{w}_k.
\label{eq:appendix_anchor}
\end{equation}
By isolating this diagonal component, we focus on the self-referential descent direction of each class, effectively filtering out gradient interference from unrelated class weights while maintaining high computational efficiency.
\newpage
\section{Additional Results}
\label{sec:append_add}

\definecolor{LightPink}{RGB}{255,238,238}

\begin{table}[ht]
\centering
\caption{\textbf{Efficiency comparison of TTA methods} on a single Nvidia A40 GPU, processing 50,000 images.}
\label{tab:efficiency}
\vskip 0.15in
\begin{small}
\resizebox{0.5\columnwidth}{!}{
\begin{tabular}{l|cccc}
\toprule
\textbf{Method} & \textbf{Forward} & \textbf{Backward} & \textbf{Time (s)} & \textbf{Memory (MB)} \\ \midrule
Source          & 50,000           & N/A               & 54.57             & 772.5               \\
TENT            & 50,000           & 50,000            & 116.00            & 5,506.7             \\
\rowcolor{LightPink} 
+STAG      & 50,000           & 50,000            & 116.25            & 5,514.0             \\

DeYO            & 84,188           & 25,961            & 151.23            & 5,956.8             \\
\rowcolor{LightPink} 
+STAG      & 85,227           & 26,811            & 156.26            & 5,965.8             \\ \bottomrule
\end{tabular}
}
\end{small}
\end{table}

\begin{algorithm}
\caption{STAG Online Adaptation}
\label{alg:stag}
\begin{algorithmic}[1]
\REQUIRE Pre-trained $f_{\bm{\phi}}$, classifier weights $\{\mathbf{w}_k\}_{k=1}^c$, $\beta_0$, decay $\gamma$, baseline loss $\mathcal{L}_{\text{base}}$
\ENSURE Adapted parameters $\theta_{\text{adapt}}$

\STATE \COMMENT{\textbf{Phase 1: Structural Anchor Derivation}}
\STATE Precompute structural anchors $\{\mathbf{a}_k\}_{k=1}^c$ using Eq.~\ref{eq:structural_anchor}

\STATE \COMMENT{\textbf{Phase 2: Online Gradient Alignment}}
\FOR{each time step $t = 0, 1, \dots$}
    \STATE Receive mini-batch $\mathcal{B}_t$ and perform forward pass
    \STATE Extract online gradients $\{\mathbf{g}(\mathbf{x})\}_{\mathbf{x} \in \mathcal{B}_t}$ via Eq.~\ref{eq:online_gradient}
    \STATE Compute alignment loss $\mathcal{L}_{\text{STAG}}(\mathcal{B}_t)$ via Eq.~\ref{eq:stag_batch_loss}
    \STATE Update $\theta_{\text{adapt}}$ by minimizing $\mathcal{L}_{\text{total}}$ (Eq.~\ref{eq:total_objective})
    \STATE $\beta_{t+1} \gets \beta_t \cdot \exp(-1/\gamma)$ \COMMENT{Adaptive decay, Eq.~\ref{eq:adaptive_decay}}
\ENDFOR
\end{algorithmic}
\end{algorithm}
\begin{figure*}[h]
    \centering
    \includegraphics[width=\textwidth]{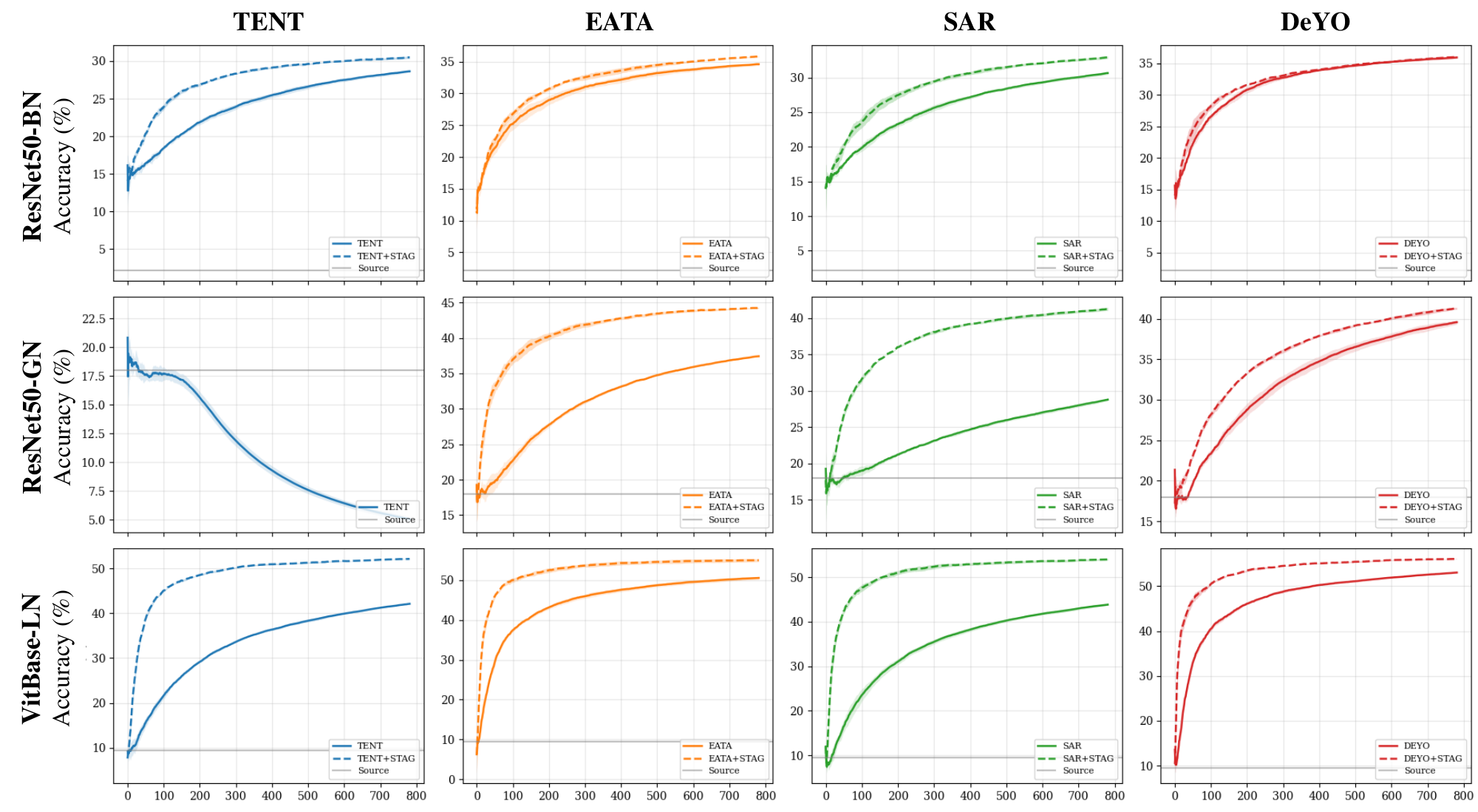}
    \caption{\textbf{Detailed accuracy running curves} for all architecture and baseline combinations on ImageNet-C Gaussian noise (severity 5). Each row represents a different model architecture, and each column denotes a specific adaptation baseline.}
    \label{fig:appendix_all_results}
\end{figure*}
\clearpage
\vspace*{\fill}
\begin{table*}[h!]
\centering
\small
\setlength{\tabcolsep}{3pt}
\renewcommand{\arraystretch}{1.15}
\caption{\textbf{Classification error rate (\%) on CIFAR-10-C in continual test-time adaptation task.} All results are evaluated on the WideResNet-28 architecture with the largest corruption severity level 5. For each baseline and its STAG-integrated version (+STAG), the superior performance is highlighted in \textbf{bold}. In cases where results are tied at the first decimal place, boldface is determined by comparing the remaining decimal digits. Values in parentheses indicate the absolute gain achieved by STAG.}
\label{tab:cifar-10-c_continual}
\resizebox{\textwidth}{!}{%
\begin{tabular}{l|ccccccccccccccc|c}
\hline
Time & \multicolumn{15}{l|}{$t \xrightarrow{\hspace{19.2cm}}$} & \\ \hline

Method & Gauss. & Shot & Impul. & Defoc. & Glass & Motion & Zoom & Snow & Frost & Fog & Brit. & Contr. & Elastic & Pixel & JPEG & Avg. \tabularnewline \hline

Source & 72.3 & 65.7 & 72.9 & 47.0 & 54.3 & 34.8 & 42.0 & 25.1 & 41.3 & 26.0 & 9.3 & 46.7 & 26.6 & 58.5 & 30.3 & 43.5 \\ \hline

BN Adapt & 28.1 & 26.1 & 36.3 & 12.8 & 35.3 & 14.2 & 12.1 & 17.3 & 17.4 & 15.2 & \textbf{8.4} & \textbf{12.7} & 23.8 & 19.6 & 27.3 & 20.4 \\ 
\rowcolor{LightPink} +STAG & $\textbf{23.1}_{\pm0.0}$ & $\textbf{19.1}_{\pm0.0}$ & $\textbf{28.5}_{\pm0.0}$ & $\textbf{12.5}_{\pm0.0}$ & $\textbf{30.8}_{\pm0.0}$ & $\textbf{13.7}_{\pm0.0}$ & $\textbf{11.7}_{\pm0.0}$ & $\textbf{16.1}_{\pm0.0}$ & $\textbf{15.6}_{\pm0.0}$ & $\textbf{14.9}_{\pm0.0}$ & $8.7_{\pm0.0}$ & $13.6_{\pm0.0}$ & $\textbf{21.5}_{\pm0.0}$ & $\textbf{17.1}_{\pm0.0}$ & $\textbf{21.9}_{\pm0.0}$ & $\textbf{17.9}_{\pm0.0}$ {\scriptsize(+2.5)} \\ \hline

TENT & 24.8 & 20.5 & 28.6 & 14.5 & 31.4 & 15.8 & 13.3 & 17.3 & 16.9 & 16.8 & 10.0 & 13.0 & 21.8 & 17.5 & 22.6 & 19.0 \\ 
\rowcolor{LightPink} +STAG & $\textbf{23.2}_{\pm0.0}$ & $\textbf{19.5}_{\pm0.0}$ & $\textbf{28.5}_{\pm0.0}$ & $\textbf{12.3}_{\pm0.0}$ & $\textbf{30.2}_{\pm0.0}$ & $\textbf{13.8}_{\pm0.0}$ & $\textbf{11.3}_{\pm0.0}$ & $\textbf{16.0}_{\pm0.0}$ & $\textbf{15.5}_{\pm0.0}$ & $\textbf{14.6}_{\pm0.0}$ & $\textbf{8.2}_{\pm0.0}$ & $\textbf{12.0}_{\pm0.1}$ & $\textbf{20.5}_{\pm0.0}$ & $\textbf{15.2}_{\pm0.1}$ & $\textbf{20.6}_{\pm0.0}$ & $\textbf{17.4}_{\pm0.0}$ {\scriptsize(+1.6)} \\ \hline

CoTTA & 24.5 & 22.0 & 26.5 & \textbf{11.8} & 27.8 & 12.4 & \textbf{10.3} & 14.9 & 14.1 & \textbf{12.5} & \textbf{7.6} & 10.7 & 18.3 & 13.5 & 17.7 & 16.3 \\ 
\rowcolor{LightPink} +STAG & $\textbf{24.1}_{\pm0.0}$ & $\textbf{21.5}_{\pm0.1}$ & $\textbf{25.8}_{\pm0.1}$ & $11.8_{\pm0.1}$ & $\textbf{27.0}_{\pm0.1}$ & $\textbf{12.3}_{\pm0.0}$ & $10.7_{\pm0.1}$ & $\textbf{14.7}_{\pm0.1}$ & $\textbf{14.1}_{\pm0.0}$ & $12.7_{\pm0.0}$ & $7.6_{\pm0.1}$ & $\textbf{10.5}_{\pm0.0}$ & $\textbf{18.3}_{\pm0.0}$ & $\textbf{13.5}_{\pm0.3}$ & $\textbf{17.5}_{\pm0.0}$ & $\textbf{16.1}_{\pm0.0}$ {\scriptsize(+0.2)} \\ \hline
    
EcoTTA {\scriptsize ($\mathrm{K}=4$)} & 25.6 & 20.9 & 28.1 & \textbf{11.6} & 31.1 & \textbf{14.3} & \textbf{11.8} & \textbf{16.8} & \textbf{15.4} & \textbf{14.0} & \textbf{8.6} & \textbf{12.9} & 20.2 & \textbf{16.2} & 20.8 & 17.9 \\ 
\rowcolor{LightPink} +STAG & $\textbf{22.5}_{\pm0.0}$ & $\textbf{19.3}_{\pm0.1}$ & $\textbf{26.2}_{\pm0.0}$ & $12.5_{\pm0.0}$ & $\textbf{29.9}_{\pm0.0}$ & $14.4_{\pm0.0}$ & $12.0_{\pm0.0}$ & $16.8_{\pm0.0}$ & $15.5_{\pm0.0}$ & $14.3_{\pm0.0}$ & $9.1_{\pm0.0}$ & $13.5_{\pm0.0}$ & $\textbf{20.2}_{\pm0.0}$ & $16.3_{\pm0.1}$ & $\textbf{20.3}_{\pm0.0}$ & $\textbf{17.5}_{\pm0.0}$ {\scriptsize(+0.4)} \\ \hline

EcoTTA {\scriptsize ($\mathrm{K}=5$)} & 25.9 & 21.1 & 28.6 & \textbf{11.5} & 31.7 & \textbf{14.2} & 11.5 & 16.9 & \textbf{15.2} & \textbf{13.6} & \textbf{9.0} & \textbf{13.0} & \textbf{20.2} & \textbf{16.3} & 20.8 & 18.0 \\ 
\rowcolor{LightPink} +STAG & $\textbf{22.7}_{\pm0.0}$ & $\textbf{19.7}_{\pm0.0}$ & $\textbf{27.4}_{\pm0.0}$ & $11.6_{\pm0.0}$ & $\textbf{30.6}_{\pm0.0}$ & $14.5_{\pm0.0}$ & $\textbf{11.5}_{\pm0.0}$ & $\textbf{16.7}_{\pm0.1}$ & $15.7_{\pm0.0}$ & $14.1_{\pm0.0}$ & $9.1_{\pm0.0}$ & $13.1_{\pm0.0}$ & $20.2_{\pm0.1}$ & $16.4_{\pm0.1}$ & $\textbf{20.3}_{\pm0.0}$ & $\textbf{17.6}_{\pm0.0}$ {\scriptsize(+0.4)} \\ \hline

\hline
\end{tabular}}
\end{table*}
\vspace{1.8em}
\begin{table*}[h!]
\centering
\small
\setlength{\tabcolsep}{3pt}
\renewcommand{\arraystretch}{1.15}
\caption{\textbf{Classification error rate (\%) on CIFAR-100-C in continual test-time adaptation task.} All results are evaluated on the WideResNet-40 architecture with the largest corruption severity level 5. For each baseline and its STAG-integrated version (+STAG), the superior performance is highlighted in \textbf{bold}. In cases where results are tied at the first decimal place, boldface is determined by comparing the remaining decimal digits. Values in parentheses indicate the absolute gain achieved by STAG.}
\label{tab:cifar-100-c_continual}
\resizebox{\textwidth}{!}{%
\begin{tabular}{l|ccccccccccccccc|c}
\hline
Time & \multicolumn{15}{l|}{$t \xrightarrow{\hspace{19.4cm}}$} & \\ \hline

Method & Gauss. & Shot & Impul. & Defoc. & Glass & Motion & Zoom & Snow & Frost & Fog & Brit. & Contr. & Elastic & Pixel & JPEG & Avg. \tabularnewline \hline

Source & 65.7 & 60.1 & 59.1 & 32.1 & 51.0 & 33.6 & 32.4 & 41.4 & 45.2 & 51.4 & 31.6 & 55.5 & 40.3 & 59.7 & 42.4 & 46.8 \\ \hline

BN Adapt & 44.3 & 44.0 & 47.3 & 32.1 & 45.9 & 32.8 & 33.0 & 38.3 & 37.9 & 45.4 & 29.9 & 36.5 & 40.6 & 36.7 & 44.1 & 39.3 \\ 
\rowcolor{LightPink} +STAG & $\textbf{39.9}_{\pm0.0}$ & $\textbf{38.2}_{\pm0.0}$ & $\textbf{42.2}_{\pm0.0}$ & $\textbf{31.2}_{\pm0.1}$ & $\textbf{43.0}_{\pm0.1}$ & $\textbf{32.7}_{\pm0.1}$ & $\textbf{31.5}_{\pm0.1}$ & $\textbf{36.1}_{\pm0.0}$ & $\textbf{35.5}_{\pm0.0}$ & $\textbf{42.5}_{\pm0.0}$ & $\textbf{29.2}_{\pm0.0}$ & $\textbf{36.2}_{\pm0.0}$ & $\textbf{38.1}_{\pm0.1}$ & $\textbf{34.3}_{\pm0.1}$ & $\textbf{40.3}_{\pm0.0}$ & $\textbf{36.7}_{\pm0.0}$ {\scriptsize(+2.6)} \\ \hline

TENT & \textbf{40.0} & \textbf{37.2} & \textbf{40.6} & 31.8 & \textbf{42.0} & 33.2 & 31.2 & 36.8 & 36.4 & 40.6 & 32.1 & 35.6 & 40.6 & 35.7 & 44.0 & 37.2 \\ 
\rowcolor{LightPink} +STAG & $40.9_{\pm0.0}$ & $38.8_{\pm0.0}$ & $42.2_{\pm0.0}$ & $\textbf{30.1}_{\pm0.0}$ & $42.3_{\pm0.0}$ & $\textbf{31.0}_{\pm0.0}$ & $\textbf{29.8}_{\pm0.0}$ & $\textbf{34.7}_{\pm0.0}$ & $\textbf{33.9}_{\pm0.0}$ & $\textbf{39.3}_{\pm0.0}$ & $\textbf{28.0}_{\pm0.0}$ & $\textbf{31.7}_{\pm0.0}$ & $\textbf{36.6}_{\pm0.0}$ & $\textbf{31.2}_{\pm0.0}$ & $\textbf{40.7}_{\pm0.0}$ & $\textbf{35.4}_{\pm0.0}$ {\scriptsize(+1.8)} \\ \hline

CoTTA & 43.5 & 41.6 & 43.7 & 32.2 & 43.6 & 32.7 & 32.2 & 38.6 & 37.5 & 46.0 & 28.9 & 38.0 & 39.3 & 33.9 & 39.3 & 38.1 \\ 
\rowcolor{LightPink} +STAG & $\textbf{43.3}_{\pm0.0}$ & $\textbf{41.2}_{\pm0.0}$ & $\textbf{42.5}_{\pm0.1}$ & $\textbf{31.7}_{\pm0.1}$ & $\textbf{42.4}_{\pm0.1}$ & $\textbf{32.5}_{\pm0.2}$ & $\textbf{31.6}_{\pm0.0}$ & $\textbf{38.0}_{\pm0.2}$ & $\textbf{36.7}_{\pm0.1}$ & $\textbf{44.9}_{\pm0.1}$ & $\textbf{28.9}_{\pm0.1}$ & $\textbf{37.2}_{\pm0.1}$ & $\textbf{38.3}_{\pm0.0}$ & $\textbf{33.5}_{\pm0.1}$ & $\textbf{38.5}_{\pm0.0}$ & $\textbf{37.4}_{\pm0.0}$ {\scriptsize(+0.7)} \\ \hline
    
EcoTTA {\scriptsize ($\mathrm{K}=4$)} & 43.5 & 40.3 & 44.7 & \textbf{31.5} & 43.8 & \textbf{32.2} & 31.1 & \textbf{35.8} & \textbf{35.0} & \textbf{40.0} & \textbf{28.4} & \textbf{34.5} & 37.6 & \textbf{32.5} & 40.8 & 36.8 \\ 
\rowcolor{LightPink} +STAG & $\textbf{41.8}_{\pm0.1}$ & $\textbf{39.2}_{\pm0.1}$ & $\textbf{43.9}_{\pm0.0}$ & $31.6_{\pm0.0}$ & $\textbf{43.7}_{\pm0.1}$ & $32.4_{\pm0.0}$ & $\textbf{31.0}_{\pm0.0}$ & $35.9_{\pm0.0}$ & $35.1_{\pm0.0}$ & $40.3_{\pm0.1}$ & $28.6_{\pm0.1}$ & $35.0_{\pm0.1}$ & $\textbf{37.5}_{\pm0.1}$ & $32.5_{\pm0.1}$ & $\textbf{40.6}_{\pm0.0}$ & $\textbf{36.6}_{\pm0.0}$ {\scriptsize(+0.2)} \\ \hline

EcoTTA {\scriptsize ($\mathrm{K}=5$)} & 44.5 & 40.9 & 44.5 & \textbf{31.1} & 43.7 & \textbf{32.1} & 31.1 & 35.9 & \textbf{34.7} & \textbf{39.9} & \textbf{28.4} & \textbf{34.3} & \textbf{37.6} & 32.4 & \textbf{40.5} & 36.8 \\ 
\rowcolor{LightPink} +STAG & $\textbf{43.0}_{\pm0.1}$ & $\textbf{40.1}_{\pm0.1}$ & $\textbf{44.3}_{\pm0.1}$ & $31.4_{\pm0.1}$ & $\textbf{43.5}_{\pm0.0}$ & $32.2_{\pm0.1}$ & $\textbf{31.1}_{\pm0.0}$ & $\textbf{35.9}_{\pm0.1}$ & $34.7_{\pm0.1}$ & $40.0_{\pm0.1}$ & $28.4_{\pm0.1}$ & $34.4_{\pm0.1}$ & $37.8_{\pm0.0}$ & $\textbf{32.4}_{\pm0.1}$ & $40.5_{\pm0.0}$ & $\textbf{36.7}_{\pm0.0}$ {\scriptsize(+0.1)} \\ \hline

\hline
\end{tabular}}
\end{table*}
\vspace{1.8em}
\begin{table*}[h!]
\centering
\small
\setlength{\tabcolsep}{3pt}
\renewcommand{\arraystretch}{1.15}
\caption{\textbf{Semantic segmentation results (mIoU in \%) on the Cityscapes-to-ACDC continual test-time adaptation task.} All results are evaluated on the Segformer-B5 architecture. For each baseline and its STAG-integrated version (+STAG), the superior performance is highlighted in \textbf{bold}. Values in parentheses indicate the absolute gain achieved by STAG.}
\label{tab:acdc_continual}
\resizebox{\textwidth}{!}{%
\begin{tabular}{l|cccc|cccc|cccc|cccc|cccc|c}
\hline
Time & \multicolumn{20}{l}{$t \xrightarrow{\hspace{25.8cm}}$} & \\ \hline
Condition & Fog & Night & Rain & Snow & Fog & Night & Rain & Snow & Fog & Night & Rain & Snow & Fog & Night & Rain & Snow & Fog & Night & Rain & Snow & cont. \\ \hline
    Round & \multicolumn{1}{l}{1} & ~ & ~ & ~ & \multicolumn{1}{l}{2} & ~ & ~ & ~ & \multicolumn{1}{l}{3} & ~ & ~ & ~ & \multicolumn{1}{l}{4} & ~ & ~ & ~ & \multicolumn{1}{l}{5} & ~ & ~ & ~ & cont. \\ \hline
    
    Source & 69.1 & 40.3 & 59.7 & 57.8 & 69.1 & 40.3 & 59.7 & 57.8 & 69.1 & 40.3 & 59.7 & 57.8 & 69.1 & 40.3 & 59.7 & 57.8 & 69.1 & 40.3 & 59.7 & 57.8 & cont. \\ 
    
    BN Adapt & 61.7 & 37.7 & 54.4 & 52.6 & 61.7 & 37.7 & 54.4 & 52.6 & 61.7 & 37.7 & 54.4 & 52.6 & 61.7 & 37.7 & 54.4 & 52.6 & 61.7 & 37.7 & 54.4 & 52.6 & cont. \\ 
    \rowcolor{LightPink} +STAG & $\textbf{62.9}_{\pm0.3}$ & $\textbf{38.7}_{\pm0.2}$ & $\textbf{55.3}_{\pm0.7}$ & $\textbf{52.8}_{\pm0.4}$ & $\textbf{62.9}_{\pm0.3}$ & $\textbf{38.7}_{\pm0.2}$ & $\textbf{55.3}_{\pm0.7}$ & $\textbf{52.8}_{\pm0.4}$ & $\textbf{62.9}_{\pm0.3}$ & $\textbf{38.7}_{\pm0.2}$ & $\textbf{55.3}_{\pm0.7}$ & $\textbf{52.8}_{\pm0.4}$ & $\textbf{62.9}_{\pm0.3}$ & $\textbf{38.7}_{\pm0.2}$ & $\textbf{55.3}_{\pm0.7}$ & $\textbf{52.8}_{\pm0.4}$ & $\textbf{62.9}_{\pm0.3}$ & $\textbf{38.7}_{\pm0.2}$ & $\textbf{55.3}_{\pm0.7}$ & $\textbf{52.8}_{\pm0.4}$ & cont. \\ 
        
    TENT & 69.0 & 40.2 & \textbf{60.0} & 57.3 & 68.3 & 39.1 & 60.0 & 56.3 & 67.5 & 37.8 & 59.6 & 55.1 & 66.5 & 36.4 & 58.7 & 54.0 & 65.7 & 35.1 & 57.8 & 53.1 & cont. \\ 
    \rowcolor{LightPink} +STAG & $\textbf{69.2}_{\pm0.0}$ & $\textbf{40.3}_{\pm0.0}$ & $59.9_{\pm0.0}$ & $\textbf{57.9}_{\pm0.0}$ & $\textbf{69.5}_{\pm0.0}$ & $\textbf{40.4}_{\pm0.0}$ & $\textbf{60.2}_{\pm0.0}$ & $\textbf{58.1}_{\pm0.0}$ & $\textbf{69.7}_{\pm0.0}$ & $\textbf{40.4}_{\pm0.0}$ & $\textbf{60.4}_{\pm0.0}$ & $\textbf{58.2}_{\pm0.0}$ & $\textbf{69.8}_{\pm0.0}$ & $\textbf{40.4}_{\pm0.0}$ & $\textbf{60.6}_{\pm0.0}$ & $\textbf{58.2}_{\pm0.0}$ & $\textbf{70.0}_{\pm0.0}$ & $\textbf{40.3}_{\pm0.0}$ & $\textbf{60.8}_{\pm0.0}$ & $\textbf{58.3}_{\pm0.0}$ & cont. \\ \hline

    Round & \multicolumn{1}{l}{6} & ~ & ~ & ~ & \multicolumn{1}{l}{7} & ~ & ~ & ~ & \multicolumn{1}{l}{8} & ~ & ~ & ~ & \multicolumn{1}{l}{9} & ~ & ~ & ~ & \multicolumn{1}{l}{10} & ~ & ~ & ~ & Avg. \\ \hline
    
    Source & 69.1 & 40.3 & 59.7 & 57.8 & 69.1 & 40.3 & 59.7 & 57.8 & 69.1 & 40.3 & 59.7 & 57.8 & 69.1 & 40.3 & 59.7 & 57.8 & 69.1 & 40.3 & 59.7 & 57.8 & 56.7 \\ 
    
    BN Adapt & 61.7 & 37.7 & 54.4 & 52.6 & 61.7 & 37.7 & 54.4 & 52.6 & 61.7 & 37.7 & 54.4 & 52.6 & 61.7 & 37.7 & 54.4 & 52.6 & 61.7 & 37.7 & 54.4 & 52.6 & 51.6 \\ 
    \rowcolor{LightPink} +STAG & $\textbf{62.9}_{\pm0.3}$ & $\textbf{38.7}_{\pm0.2}$ & $\textbf{55.3}_{\pm0.7}$ & $\textbf{52.8}_{\pm0.4}$ & $\textbf{62.9}_{\pm0.3}$ & $\textbf{38.7}_{\pm0.2}$ & $\textbf{55.3}_{\pm0.7}$ & $\textbf{52.8}_{\pm0.4}$ & $\textbf{62.9}_{\pm0.3}$ & $\textbf{38.7}_{\pm0.2}$ & $\textbf{55.3}_{\pm0.7}$ & $\textbf{52.8}_{\pm0.4}$ & $\textbf{62.9}_{\pm0.3}$ & $\textbf{38.7}_{\pm0.2}$ & $\textbf{55.3}_{\pm0.7}$ & $\textbf{52.8}_{\pm0.4}$ & $\textbf{62.9}_{\pm0.3}$ & $\textbf{38.7}_{\pm0.2}$ & $\textbf{55.3}_{\pm0.7}$ & $\textbf{52.8}_{\pm0.4}$ & $\textbf{52.4}_{\pm0.2}$ {\scriptsize(+0.8)} \\
    
    TENT & 65.0 & 34.0 & 56.6 & 52.2 & 64.3 & 33.0 & 55.5 & 51.2 & 63.5 & 31.9 & 54.2 & 50.2 & 62.7 & 30.9 & 53.2 & 49.2 & 62.1 & 30.1 & 52.2 & 48.3 & 52.4 \\ 
    \rowcolor{LightPink} +STAG & $\textbf{70.0}_{\pm0.0}$ & $\textbf{40.2}_{\pm0.0}$ & $\textbf{61.0}_{\pm0.0}$ & $\textbf{58.3}_{\pm0.0}$ & $\textbf{70.1}_{\pm0.0}$ & $\textbf{40.2}_{\pm0.0}$ & $\textbf{61.1}_{\pm0.0}$ & $\textbf{58.3}_{\pm0.0}$ & $\textbf{70.1}_{\pm0.0}$ & $\textbf{40.1}_{\pm0.0}$ & $\textbf{61.2}_{\pm0.0}$ & $\textbf{58.2}_{\pm0.0}$ & $\textbf{70.0}_{\pm0.0}$ & $\textbf{39.9}_{\pm0.0}$ & $\textbf{61.2}_{\pm0.0}$ & $\textbf{58.1}_{\pm0.0}$ & $\textbf{70.0}_{\pm0.0}$ & $\textbf{39.7}_{\pm0.0}$ & $\textbf{61.2}_{\pm0.0}$ & $\textbf{58.0}_{\pm0.0}$ & $\textbf{57.2}_{\pm0.0}$ {\scriptsize(+4.8)} \\ \hline
        
\end{tabular}}
\end{table*}
\vspace*{\fill}
\clearpage
\begin{table}[ht]
\centering
\caption{Hyper-parameter configurations ($\beta_0$ and $\gamma$) for all experimental settings. We report the initial balancing coefficient $\beta_0$ and the decay scale $\gamma$ for each combination of task, scenario, architecture, and baseline method.}
\label{tab:appendix_hp}
\footnotesize
\begin{tabular}{lllllrr}
\toprule
\textbf{Task} & \textbf{Scenario} & \textbf{Dataset} & \textbf{Model} & \textbf{Baseline Method} & $\beta_0$ & $\gamma$ \\
\midrule
\multirow{41}{*}{Classification} & \multirow{11}{*}{Mild} & \multirow{11}{*}{ImageNet-C} & \multirow{4}{*}{ResNet50-BN} & TENT & 100 & 100 \\
 & & & & EATA & 25 & 150 \\
 & & & & SAR & 50 & 150 \\
 & & & & DeYO & 12 & 100 \\
\cmidrule{4-7}
 & & & \multirow{3}{*}{ResNet50-GN} & EATA & 250 & 100 \\
 & & & & SAR & 150 & 100 \\
 & & & & DeYO & 25 & 50 \\
\cmidrule{4-7}
 & & & \multirow{4}{*}{VitBase-LN} & TENT & 300 & 50 \\
 & & & & EATA & 100 & 100 \\
 & & & & SAR & 500 & 50 \\
 & & & & DeYO & 250 & 50 \\
\cmidrule{2-7}
 & \multirow{7}{*}{Label Shifts} & \multirow{7}{*}{ImageNet-C} & \multirow{3}{*}{ResNet50-GN} & EATA & 125 & 100 \\
 & & & & SAR & 100 & 70 \\
 & & & & DeYO & 25 & 50 \\
\cmidrule{4-7}
 & & & \multirow{4}{*}{VitBase-LN} & TENT & 70 & 50 \\
 & & & & EATA & 100 & 100 \\
 & & & & SAR & 250 & 100 \\
 & & & & DeYO & 25 & 50 \\
\cmidrule{2-7}
 & \multirow{6}{*}{Mix Shifts} & \multirow{6}{*}{ImageNet-C} & \multirow{3}{*}{ResNet50-GN} & EATA & 500 & 100 \\
 & & & & SAR & 200 & 1000 \\
 & & & & DeYO & 25 & 750 \\
\cmidrule{4-7}
 & & & \multirow{3}{*}{VitBase-LN} & EATA & 100 & 1500 \\
 & & & & SAR & 500 & 1000 \\
 & & & & DeYO & 50 & 1000 \\
\cmidrule{2-7}
 & \multirow{7}{*}{Batch Size 1} & \multirow{7}{*}{ImageNet-C} & \multirow{3}{*}{ResNet50-GN} & EATA & 500 & 10000 \\
 & & & & SAR & 500 & 10000 \\
 & & & & DeYO & 25 & 10000 \\
\cmidrule{4-7}
 & & & \multirow{4}{*}{VitBase-LN} & TENT & 300 & 10000 \\
 & & & & EATA & 500 & 10000 \\
 & & & & SAR & 1000 & 10000 \\
 & & & & DeYO & 250 & 10000 \\
\cmidrule{2-7}
 & \multirow{10}{*}{Continual} & \multirow{5}{*}{CIFAR-10-C} & \multirow{5}{*}{WideResNet-28} & BN Stats Adapt & 90 & 15 \\
 & & & & TENT & 100 & 10 \\
 & & & & CoTTA & 5 & 10 \\
 & & & & EcoTTA ($K=4$) & 30 & 60 \\
 & & & & EcoTTA ($K=5$) & 100 & 10 \\
\cmidrule{3-7}
 & & \multirow{5}{*}{CIFAR-100-C} & \multirow{5}{*}{WideResNet-40} & BN Stats Adapt & 70 & 15 \\
 & & & & TENT & 500 & 5 \\
 & & & & CoTTA & 10 & 400 \\
 & & & & EcoTTA ($K=4$) & 90 & 35 \\
 & & & & EcoTTA ($K=5$) & 120 & 10 \\
\midrule
\multirow{2}{*}{Segmentation} & \multirow{2}{*}{Continual} & \multirow{2}{*}{Cityscapes-ACDC} & \multirow{2}{*}{Segformer-B5} & BN Stats Adapt & 250 & 25 \\
 & & & & TENT & 500 & 35 \\
\bottomrule
\end{tabular}
\end{table}


\end{document}